\documentclass[draftcls,11pt,onecolumn]{IEEEtran}
\usepackage{amsmath,amssymb,latexsym,cite}
\usepackage{pstricks,graphicx}
\usepackage[caption=false]{subfig}
\usepackage{xcolor}
\usepackage{algorithm,algpseudocode}
\def\mbf{\mathbf}
\def\mc{\mathcal}

\newcommand{\expe}[1]{ \mathbb{E}\left[ #1 \right] }
\newcommand{\E}{\mathbb{E}}
\newcommand{\prob}{\mathbb{P}}
\newcommand{\norm}[1]{ \left\| #1 \right\| }

\def\defeq{ \stackrel{\Delta}{=} }
\newcommand{\field}[1]{\mathbb{#1}}
\newcommand{\R}{\field{R}}
\newcommand{\argmax}{\mathop{\mathrm{argmax}}}

\newcommand{\x}{\mbf{x}}
\newcommand{\Xmap}{{\mbf{x}^*}}
\newcommand{\Xrmap}{{\mbf{x}^{*}}}

\newcommand{\cX}{\mc{X}}
\newcommand{\dompot}{\mathrm{Dom}(\theta)}
\newcommand{\pot}{\theta}
\newcommand{\gumv}{\gamma}
\newcommand{\gum}[1]{\gumv(#1)}

\newcommand{\gumset}{\gamma}

\def\Ent{\mathop{\mathrm{Ent}}\nolimits}
\def\Var{\mathop{\mathrm{Var}}\nolimits}
\newcommand{\cvxf}{Q}  %
\newcommand{\lcdens}{q}  %
\newcommand{\h}{f}  %
\newcommand{\mvh}{F} %

\newtheorem{theorem}{Theorem}
\newtheorem{lemma}{Lemma}

\newtheorem{corollary}{Corollary}

\renewcommand{\[}{\begin{eqnarray}}
\renewcommand{\]}{\end{eqnarray}}

\title{High Dimensional Inference with Random Maximum A-Posteriori Perturbations}
\author{
Tamir Hazan,
Francesco Orabona,
Anand~D.~Sarwate~\IEEEmembership{Senior Member,~IEEE,} 
Subhransu~Maji 
and~Tommi Jaakkola
\thanks{Manuscript received February 10, 2016;
revised November 4, 2016; accepted xxxxxxxxxx  }%
\thanks{T. Hazan is with the Faculty of Industrial Engineering \& Management, Technion - Israel Institute of Technology, Technion City, Haifa 32000, Israel (e-mail: \texttt{tamir.hazan@technion.ac.il}).
F. Orabona is with the Department of Computer Science, Stony Brook University, Stony Brook, NY 11794-2424. This work was done in part when he was with Yahoo Labs, 229 W 43rd St., New York, NY 10036, USA (e-mail: \texttt{francesco@orabona.com}).
A.D. Sarwate is with the Department of Electrical and Computer Engineering, Rutgers, The State University of New Jersey, 94 Brett Road, Piscataway NJ 08854, USA (e-mail: \texttt{asarwate@ece.rutgers.edu}).  
S. Maji is with the College of Information and Computer Sciences, University of Massachusetts Amherst, 140 Governors Drive, MA 01003-9264, USA (e-mail: \texttt{smaji@cs.umass.edu}).  %
T. Jaakkola is with the Department of Electrical Engineering and Computer Science, Massachusetts Institute of Technology, 77 Massachusetts Avenue, Cambridge, MA 02139, USA (e-mail: \texttt{tommi@csail.mit.edu}).}
\thanks{Preliminary versions of these results appeared in conference proceedings~\cite{HazanJ:12icml,Hazan13-gibbs,Orabona14,Maji14}.}
}

\begin{document}

\maketitle

\begin{abstract}
This paper presents a new approach, called perturb-max, for high-dimensional statistical inference that is based on applying random perturbations followed by optimization. This framework injects randomness to maximum a-posteriori (MAP) predictors by randomly perturbing the potential function for the input. A classic result from extreme value statistics asserts that perturb-max operations generate unbiased samples from the Gibbs distribution using high-dimensional perturbations.  Unfortunately, the computational cost of generating so many high-dimensional random variables can be prohibitive. However, when the perturbations are of low dimension, sampling the perturb-max prediction is as efficient as MAP optimization. This paper shows that the expected value of perturb-max inference with low dimensional perturbations can be used sequentially to generate unbiased samples from the Gibbs distribution. Furthermore the expected value of the maximal perturbations is a natural bound on the entropy of such perturb-max models. A measure concentration result for perturb-max values shows that the deviation of their sampled average from its expectation decays exponentially in the number of samples, allowing effective approximation of the expectation.
\end{abstract}


\section{Introduction \label{sec:intro}}

Modern machine learning tasks in computer vision, natural language processing, and computational biology involve inference in high-dimensional  models.  Examples include scene understanding~\cite{Felz11}, parsing~\cite{Koo10}, and protein design~\cite{Sontag-uai08}. In these settings, inference involves finding a likely assignment (or equivalently, structure) that fits the data: objects in images, parsers in sentences, or molecular configurations in proteins. Each structure corresponds to an assignment of values to random variables and the preference of a structure is based on defining potential functions that account for interactions over these variables. Given the observed data, these preferences yield a \textit{posterior probability distribution} on assignments called the Gibbs distribution. The probability of an assignment is proportional to the exponential of the potential function value.  High dimensional models that are commonly used in contemporary machine learning often incorporate local potential functions on the variables of the model that are derived from the data (signal) as well as higher order potential functions that account for interactions between the model variables and derived from domain-specific knowledge (coupling).  The resulting posterior probability landscape is often ``ragged''; in such landscapes Markov chain Monte Carlo (MCMC) approaches to sampling from the Gibbs distribution may become prohibitively expensive~\cite{Jerrum93, Goldberg07, Goldberg12}.  By contrast, when no data terms (local potential functions) exist, MCMC approaches can be quite successful. These methods include Gibbs sampling~\cite{Geman84}, Metropolis-Hastings~\cite{Hastings70}, or Swendsen-Wang~\cite{Wang87}.

\begin{figure}[t]
\begin{center}
\includegraphics[width=2.0in]{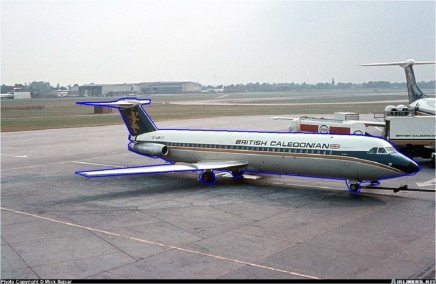} \hspace{0.2cm}
\includegraphics[width=2.0in]{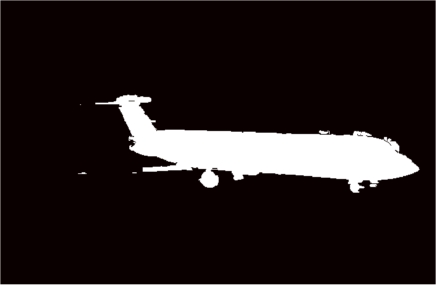} \hspace{0.2cm}
\includegraphics[width=2.0in]{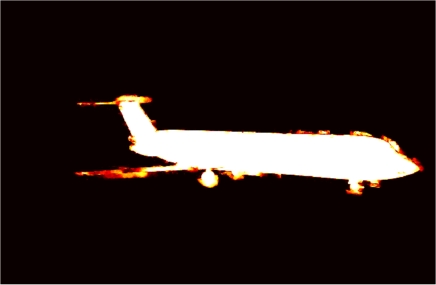}
\end{center}
\caption{{\label{fig:plane} \small Comparing MAP inference and perturbation models. A segmentation is modeled by $\x=(x_1,x_2,\ldots,x_n)$ where $n$ is the number of pixels and $x_i \in \{0,1\}$ is a discrete label relating a pixel to foreground ($x_i = 1$) or background ($x_i = 0$). $\pot(\x)$ is the (super-modular) score of each segmentation. Left: original image along with the annotated boundary. Middle: the MAP segmentation $\argmax_{\x} \pot(\x)$ recovered by the graph-cuts optimization algorithm using a region inside the boundary as seed pixels\cite{Boykov01}. Note that the ``optimal'' solution is inaccurate because thin and long objects (wings) are labeled incorrectly. Right: The marginal probabilities of the perturb-max model estimated using 20 samples (random perturbations of $\pot(\x)$ followed by executing graph-cuts). The information about the wings is recovered by these samples. Estimating the marginal probabilities of the corresponding Gibbs distribution by MCMC sampling is slow in practice and provably hard in theory~\cite{Hazan13-gibbs,Goldberg07}}.}
\end{figure}

An alternative to sampling from the Gibbs distribution is to look for the \textit{maximum a posteriori probability} (MAP) assignment. Substantial effort has gone into developing optimization algorithms for recovering MAP assignments by exploiting domain-specific structural restrictions~\cite{Eisner96,Boykov01,Kolmogorov-pami06,Gurobi,Felz11,Swoboda13} or by linear programming relaxations~\cite{Wainwright05-map,Weiss-uai07,Sontag-uai08,Werner08,Peng12}.
MAP inference is nevertheless limiting when there are a number of alternative likely assignments. Such alternatives arise either from inherent ambiguities (e.g., in image segmentation or text analysis) or due to the use of computationally/representationally limited potential functions (e.g., super-modularity) aliasing alternative assignments to have similar scores. For an example, see Figure \ref{fig:plane}. 

Recently, several works have leveraged the current efficiency of MAP solvers to build (approximate) samplers for the Gibbs distribution, thereby avoiding the computational burden of MCMC methods~\cite{Papandreou11,Tarlow12,Hazan13-gibbs,Ermon13-icml,Ermon13-uai,Ermon13-nips,Ermon14-icml,Maddison14,Papandreou14,Gane14,Keshet11,Kalai05}.  
These works have shown that one can represent the Gibbs distribution by calculating the MAP assignment of a \textit{randomly perturbed potential function}, whenever the perturbations follow the Gumbel distribution~\cite{Papandreou11,Tarlow12}. Unfortunately the total number of assignment (or structures), and consequently the total number of random perturbations, is exponential in the structure's dimension. We call this a \textit{perturb-max} approach.

In this work, we perform high dimensional inference tasks using the expected value of perturb-max programs that are restricted to low dimensional perturbations. In this setting, the number of random perturbations is linear is the assignment's dimension and as a result statistical inference is as fast as computing the MAP assignment, as illustrated in Figure \ref{fig:plane}. 
We also provide measure concentration inequalities that show the expected perturb-max value can be estimated with high probability using only a few random samples. This work simplifies and extends our preliminary results~\cite{HazanJ:12icml,Hazan13-gibbs,Maji14,Orabona14}. %

We begin by introducing the setting of high dimensional inference as well as the necessary background in extreme value statistics in Section \ref{sec:inference}. Subsequently, we develop high dimensional inference algorithms that rely on the expected MAP value of randomly perturbed potential functions, while using only low dimensional perturbations. In Section \ref{sec:unbiased} we propose a novel sampling algorithm and in Section \ref{sec:entropy} we derive bounds on the entropy that may be of independent interest. Finally, we show that the expected value of the perturb-max value can be estimated efficiently despite the unboundedness of the perturbations. To show this we must prove new measure concentration results for the Gumbel distribution. In particular, in Section \ref{sec:mc} we prove new Poincar\'{e} and modified log-Sobolev inequalities for (non-strictly) log-concave distributions.   

\section{Inference and random perturbations \label{sec:inference}}

We first describe the high dimensional statistical inference problems that motivate this work. These involve defining the potential function, the Gibbs distribution, and its entropy. Further background can be found in standard texts on graphical models~\cite{Wainwright08}. We will then describe the MAP inference problem and describe how to use extreme value statistics to perform statistical inference while recovering the maximal assignment of randomly perturbed potential functions~\cite{Kotz00}\cite[pp.159--61]{David03}. To do this, we apply random perturbations to the potential function and use MAP solvers to produce a solution to the perturbed problem.

\subsection{High dimensional models, inference and extreme value statistics}

Statistical inference for high dimensional problems involves reasoning about the states of discrete variables whose configurations (assignments of values) describe discrete structures. 
Suppose that our model has $n$ variables $\x = (x_1, x_2, \ldots, x_n)$ where each $x_i$ takes values in a discrete set $\cX_i$.  Let $\cX = \cX_1 \times \cX_2 \times \cdots \times \cX_n$ so that $\x \in \cX$. Let $\dompot \subseteq \cX$ be a subset of possible configurations and $\pot : \cX \to \mathbb{R}$ be a potential function that gives a score to an assignment or structure $\x$. For convenience we define $\pot(\x) = -\infty$ for $\x \notin \dompot$.  The potential function induces a probability distribution on configurations $\x = (x_1,x_2, \ldots,x_n)$ via the Gibbs distribution: 
\begin{align}
\label{eq:gibbs}
p(\x) \defeq \frac{1}{Z(\theta)} \exp(\pot(\x)) \qquad \mbox{where} \qquad Z(\pot) \defeq \sum_{\x \in \cX} \exp(\pot(\x)). 
\end{align}
The normalization constant $Z(\theta)$ is called the partition function. Sampling from the Gibbs distribution is often difficult because the partition function involves exponentially many terms (equal to the number of discrete structures in $\mc{X}$). Computing the partition function is $\#P$-hard in general (e.g., Valiant~\cite{Valiant79}).

\subsection{MAP inference}

In practical inference tasks, the Gibbs distribution is constructed given observed data. Thus we call its maximizing assignment the maximum a-posteriori (MAP) prediction. We express the MAP inference problem as maximizing $p(\x)$, which is defined in Equation  \eqref{eq:gibbs}. Since the exponent is a monotone function, maximizing $p(\x)$ is equivalent to maximizing $\pot(\x)$ and MAP prediction amounts to finding
	\begin{align}
	\Xmap =  \argmax_{\x \in \cX} \pot(\x). \label{eq:M}
\end{align}
Methods for performing the optimization in \eqref{eq:M} for high dimensional potential functions have been extensively researched in the last decade~\cite{Boykov01,Sontag-uai08,Gurobi,Felz11,Swoboda13}. These have been useful in many cases of practical interest in computer vision, such as foreground-background image segmentation with supermodular potential functions (e.g.,~\cite{Kolmogorov04}), parsing and tagging (e.g.,~\cite{Koo10, Rush10}), branch and bound for  scene understanding and pose estimation~\cite{Schwing12-eccv, Sun12}, and dynamic programming predictions for outdoor scene understanding~\cite{Felzenszwalb10}. Although the run-time of these solvers can be exponential in the number of variables, they are often surprisingly effective in practice,~\cite{Wainwright05-map,Globerson07nips,Sontag-uai08,Sontag-nips08,Peng12}. 

\subsection{Inference and extreme value statistics}

Although MAP prediction is NP-hard in general, it is often simpler than sampling from the Gibbs distribution. Nevertheless, usually there are several values of $\x$ whose scores $\pot(\x)$ are close to $\pot(\Xmap)$ and we would like to sample these structures (see Figure \ref{fig:plane}). From such samples it is possible to estimate the amount of uncertainty in these models. A standard uncertainty measure is the entropy function: 
\begin{align}
\label{eq:entropy}
H(p) &= - \sum_{\x \in \cX} p(\x) \log p(\x).
\end{align}
Sampling methods for posterior distributions often resort to MCMC algorithms that converge slowly in many practical settings~\cite{Jerrum93, Goldberg07, Goldberg12}. 

An alternative approach to drawing unbiased samples from the Gibbs distribution is by randomly perturbing the potential function and solving the perturbed MAP problem.  The ``perturb-max'' approach adds a random function $\gumv : \cX \to \mathbb{R}$ to the potential function in \eqref{eq:gibbs} and solves the resulting MAP problem:
	\begin{align}
	\Xrmap =  \argmax_{\x \in \cX} \left\{ \pot(\x) + \gum{\x} \right\}, \label{eq:Rmap} 
	\end{align}
where $\gum{\x}$ is a random function on $\cX$.
The simplest approach to designing a perturbation function is to associate an independent and identically distributed (i.i.d.) random variable $\gum{\x}$ for each $\x \in \cX$.  In this case, the distribution of the perturb-max value $\pot(\x) + \gamma(\x)$ has an analytic form. To verify this observation we denote by $F(t) = \mathbb{P}(\gamma(\x) \le t)$ the cumulative distribution function of $\gamma(\x)$.
The independence of $\gum{\x}$ across $\x \in \cX$ implies that
	\begin{align}
	\prob_{\gumv}\left(\max_{\x \in \cX} \left\{ \pot(\x) + \gum{\x} \right\} \le t \right) 
	&= \prob_{\gumv}\left( \forall \x \in \cX:  \left\{ \pot(\x) + \gum{\x} \right\} \le t \right) \\
	&= \prob_{\gumv}\left( \forall \x \in \cX:  \left\{ \pot(\x) + \gum{\x} \right\} \le t \right) \\
	&= \prod_{\x \in \cX} F(t-\pot(\x)).
	\end{align}
Unfortunately, the product of cumulative distribution functions is not usually a simple distribution. 

The Gumbel, Fr\'echet, and Weibull distributions, used in extremal statistics, are max-stable distributions: the product $\prod_{\x \in \cX} F(t-\pot(\x))$ can be described by their own cumulative distribution function $F(\cdot)$~\cite{Fisher28,Gnedenko43,Gumbel54}. In this work we focus on the Gumbel distribution with zero mean, which is described by a doubly exponential cumulative distribution function
	\begin{align}
	G( t ) = \exp( - \exp( - (t + c))),
	\label{eq:gumbelcdf}
	\end{align}
where $c \approx 0.5772$ is the Euler-Mascheroni constant. 
Throughout our work we use the max-stability of the Gumbel distribution as described in the following theorem.

\begin{theorem}[Max-stability of Gumbel perturbations~\cite{Fisher28,Gnedenko43,Gumbel54}]
\label{theorem:z}
Let $\gumset = \{ \gum{\x} : \x \in \cX \}$ be a collection of i.i.d. Gumbel random variables whose cumulative distribution function is given by $G(t) = \prob( \gamma(\x) \le t ) = \exp( - \exp( - (t + c)))$.  Then the random variable $\max_{\x \in \cX} \left\{ \pot(\x) + \gum{\x} \right\}$ is distributed according to the Gumbel distribution whose mean is the log-partition function $ \log Z(\pot)$.
\end{theorem}
\begin{IEEEproof}
The proof is known, but we include it for completeness.
By the independence assumption,
	\begin{align*}
	\prob_{\gumset}\left( \max_{\x \in \cX} \{\pot(\x) + \gum{\x}\} \le t \right) = \prod_{x \in X}\prob_{\gamma(\x)}\left( \pot(\x) + \gum{\x}  \le t \right).
	\end{align*}
The random variable $\pot(\x) + \gum{\x}$ follows the Gumbel distribution with mean $\pot(\x)$. Therefore 
	\begin{align*}
	\prob_{\gamma(x)}\left( \pot(\x) + \gum{\x}  \le t \right) = G( t - \pot(\x)).
	\end{align*}
Lastly, the double exponential form of the Gumbel distribution yields the result: 
	\begin{align*}
	\prod_{x \in X} G( t - \pot(\x)) &= \exp\left( - \sum_{\x \in \cX} \exp\left( - (t - \pot(\x) + c) \right) \right) \\
	&= \exp\left( - \exp( - (t + c - \log Z(\pot)) \right) \\
	&= G(t - \log Z(\pot)).
	\end{align*}
 \end{IEEEproof}
 
We can use the log-partition function to recover the moments of the Gibbs distribution. Thus the log-partition function characterizes the stability of the randomized MAP predictor $\Xrmap$ in \eqref{eq:Rmap}.

\begin{corollary}[Sampling from perturb-max models~\cite{Luce59,Ben85,McFadden73}]
\label{corollary:p}
Under the conditions of Theorem \ref{theorem:z} the Gibbs distribution measures the stability of the perturb-max argument. That is, for all $\hat \x$,
\begin{align}
\frac{\exp(\pot(\hat{\x}))}{Z(\pot)} &= \prob_{\gumset} \left(
	\hat{\x} = \argmax_{\x \in \cX} \left\{ \pot(\x) + \gum{\x} \right\}
	\right), 
	\label{eq:gumbel-p}
\end{align}
\end{corollary}

\begin{IEEEproof}
From Theorem \ref{theorem:z}, we have $\log Z(\pot) = \E_{\gumv}[\max_{\x \in \cX} \left\{ \pot(\x) + \gum{\x} \right\}]$, so we can take the derivative with respect to some $\pot(\hat{\x})$. We note that by differentiating the left hand side we get the Gibbs distribution:
	\begin{align*}
	\frac{\partial \log Z(\pot) }{ \partial \pot(\hat{\x}) } 
	= \frac{\exp(\pot(\hat{\x}))}{Z(\pot)}.
	\end{align*}
Differentiating the right hand side is slightly more involved. First, we can differentiate under the integral sign (cf.~\cite{Folland13}) so 
	\begin{align*}
	\frac{\partial}{\partial \pot(\hat{\x})}  
	\int_{\mathbb{R}^{|\cX|}} \max_{ \x \in \cX } \left\{ \pot(\x) + \gum{\x} \right\} d\gumv
	 &=  \int_{\mathbb{R}^{|\cX|}} \frac{\partial }{\partial \pot(\hat{\x})} \max_{ \x \in \cX } \left\{ \pot(\x) + \gum{\x} \right\}  d\gumv.
	 \end{align*}
The (sub)gradient of the max-function is the indicator function (an application of Danskin's Theorem~\cite{Bertsekas03}): 
	\begin{align*}
	\frac{\partial}{\partial \pot(\hat{\x})} \max_{ \x \in \cX } \left\{ \pot(\x) + \gum{\x} \right\} &= \mbf{1}\left( \hat{\x} = \argmax_{ \x \in \cX } \left\{ \pot(\x) + \gum{\x} \right\} \right).
	\end{align*}
The corollary then follows by applying the expectation to both sides of the last equation.
\end{IEEEproof}

An alternative proof of the preceding corollary can be given by considering the probability density function $g(t) = G'(t)$ of the Gumbel distribution. This proof consists of two steps. First, the probability that $\hat{\x}$ maximizes $\pot(\x)+ \gamma(\x)$ is $\int g(t-\pot(\hat{\x})) \prod_{\x \ne \hat{\x}} G(t-\pot(\x)) dt$. Second, $g(t-\pot(\hat{\x})) = \exp(\pot(\hat{\x})) \cdot \exp(-(t+c)) G(t-\pot(\hat{\x}))$. Thus, the probability that $\hat{\x}$ maximizes $\pot(\x)+ \gamma(\x)$ is proportional to $\exp( \pot(\hat{\x}))$, i.e., it is the Gibbs distribution.

We can also use the random MAP perturbation to estimate the entropy of the Gibbs distribution.

\begin{corollary}
\label{corollary:entropy}
Let $p(x)$ be the Gibbs distribution, defined in \eqref{eq:entropy} and let $\Xrmap$ be given by \eqref{eq:Rmap}. Under the conditions of Theorem \ref{theorem:z},
\begin{align*}
H(p) &= \E_{\gumv}\left[ \gum{\Xrmap}  \right].
\end{align*}
\end{corollary}

\begin{IEEEproof}
The proof consists of evaluating the entropy in Equation \eqref{eq:entropy} and using Theorem \ref{theorem:z} to replace $\log Z(\pot)$ with $\E_{\gumv}[ \pot(\Xrmap) + \gum{\Xrmap} ]$. Formally, 
\begin{align*}
H(p) &= -\sum_{\x \in \cX} p(\x) \pot(\x) + \E_{\gumv}[ \pot(\Xrmap) + \gum{\Xrmap} ] \nonumber \\
&= -\sum_{\x \in \cX} p(\x) \pot(\x) + \sum_{\x \in \cX} \pot(\x) \prob_{\gumv}( \Xrmap = \x ) + \E_{\gumv}[ \gum{\Xrmap} ] \nonumber \\
&= \E_{\gumv}[ \gum{\Xrmap} ], \nonumber
\end{align*}
where in the last line we used Corollary \ref{corollary:p}, which says $\prob_{\gumv}( \Xrmap = \x ) = p(\x)$. 
\end{IEEEproof}

A direct proof of the preceding corollary can be given by showing that $\E_{\gumv}\left[ \gum{\Xrmap} \cdot 1[\hat{\x} = \Xrmap] \right] = -p(\hat{\x}) \log p(\hat{\x})$ while the entropy is then attained by summing over all $\hat{\x}$, since $\sum_{\hat{\x} \in \cX} 1[\hat{\x} = \Xrmap] = 1$. To establish this equality we note that 
	\begin{align*}
	\E_{\gumv}\left[ \gum{\Xrmap} \cdot 1[\hat{\x} = \Xrmap] \right] 
	&= \int (t-\pot(\hat{\x})) g(t-\pot(\hat{\x})) \prod_{\x \ne \hat{\x}} G(t-\pot(\x)) dt.
	\end{align*}
Using the relation between $g(t)$ and $G(t)$ and the fact that $\prod_{x \in X} G(t-\pot(x)) = G(t-\log Z(\pot))$ while changing the integration variable to $\hat t =  t-\pot(\hat{\x})$ we can rephrase this quantity as $\int t \exp(-(c+t)) G(t  + \log p(\hat{\x})) dt$. Again by using the relation between $g(t+\log p(\hat{\x}))$ and $G(t + \log p(\hat{\x}))$ we derive that $\E_{\gumv}\left[ \gum{\Xrmap} \cdot 1[\hat{\x} = \Xrmap] \right] = p(\hat{\x}) \int t g(t  + \log p(\hat{\x})) dt$ while the integral is now the mean of a Gumbel random variable with expected value of $-\log p(\hat{\x})$.

The preceding derivations show that perturbing the potential function $\pot(\x)$ and then finding the MAP estimate $\Xrmap$ of the perturbed Gibbs distribution allows us to perform many core tasks for high-dimensional statistical inference by using i.i.d. Gumbel perturbations.  %
The distribution of $\Xrmap$ is $p(\x)$, its expected maximum value is the log-partition function, and the expected maximizing perturbation is the entropy of $p(\x)$.  While theoretically appealing, these derivations are computationally intractable when dealing with high-dimensional structures. These derivations involve generating high-dimensional perturbations, namely $|\cX|$ random variables in the image of $\gum{\cdot}$, one for each assignment in $\cX = \cX_1 \times \cdots \cX_n$, which grows exponentially with $n$. The goal of this paper is to apply high-dimensional inference using max-solvers that involve a low-dimensional perturbation term. More specifically, we wish to involve only a linear (in $n$) number of random variables.

%
\section{Low-dimensional perturbations}

We now turn towards making the perturb-max framework more practical. The log-partition function $\log Z(\pot)$ (c.f. Theorem \ref{theorem:game}) is the key quantity to understand: its gradient is the Gibbs distribution and the entropy is its Fenchel dual. It is well-known that computing $Z(\pot)$ for high-dimensional models is challenging because of the exponential size of $\cX$. This complexity carries over to the perturb-max approach to estimating the log-partition function, which also involves generating an exponential number of Gumbel random variables.  In this section we show that the log-partition function can be computed using low-dimensional perturbations in a sequence of expected max-value computations. This will give us some insight on performing high dimensional inference using low dimensional perturbations. In what follows we will use the notation $\x_{i}^{j}$ to refer to the tuple $(x_i, x_{i+1}, \ldots, x_{j})$ for $i < j$, with $\mbf{x} = \x_{1}^{n}$.

The partition function has a self-reducible form. That is, we can compute it iteratively while computing partial partition functions of lower dimensions: 
	\begin{align}
	Z(\pot) = \sum_{x_1} \sum_{x_2} \cdots \sum_{x_n} \exp (\pot(x_1,x_2,\ldots,x_n)). 
	\label{eq:sum_partition}
	\end{align}	
For example, the partition function is the sum, over $x_1$, of partial partition functions $\sum_{x_2, \ldots, x_n} \exp (\pot(\x))$. Fixing $x_1,x_2,\ldots,x_i$, the remaining summations are partial partition functions $\sum_{x_{i+1},\ldots,x_n} \exp (\pot(\x))$. With this in mind, we can compute each partial partition function using Theorem \ref{theorem:z} but with low-dimensional perturbations for each partial partition. 

\begin{theorem}
\label{theorem:game}
Let $\{\gamma_i(x_i)\}_{x_i\in \cX_i, i=1,\ldots,n}$, be a collection of independent and identically distributed (i.i.d.) random variables following the Gumbel distribution, defined in Theorem \ref{theorem:z}. Define $\gamma_i = \{\gamma_i(x_i)\}_{x_i\in \cX_i}$. Then 
	\begin{align}
	\log Z = \E_{\gamma_1} \max_{x_1} \cdots \E_{\gamma_n} \max_{x_n}  \left\{ \pot(\x) + \sum_{i=1}^n \gamma_i(x_i) \right\}. 
	\label{eq:Emax}
	\end{align}
\end{theorem}

\begin{IEEEproof}
The result follows from applying Theorem \ref{theorem:z} iteratively. Let $\pot_n(\x_1^n) = \pot(\x_1^n)$ and define
	\begin{align*}
	\pot_{i-1}(\x_1^{i-1}) 
		= \E_{\gamma_i} \max_{x_i}  \{ \pot_i(\x_1^i) + \gamma_i(x_i) \} 
		\qquad i = 2, 3, \ldots, n
	\end{align*}
If we think of $\x_1^{i-1}$ as fixed and apply Theorem \ref{theorem:z} to $\pot_i(\x_1^{i-1},x_i)$, we see that from \eqref{eq:sum_partition}, 
	\begin{align*}
	\pot_{i-1}(\x_1^{i-1}) 
		= \log \sum_{x_i} \exp( \pot_i( \x_{1}^i ) ).
	\end{align*}
Applying this for $i = n$ to $i = 2$, we obtain \eqref{eq:Emax}.
\end{IEEEproof}

The computational complexity of the alternating procedure in \eqref{eq:Emax} is still exponential in $n$. For example, the innermost iteration $\pot_{n-1}(\x_1^{n-1}) = \E_{\gamma_n} \max_{x_n} \{ \pot_n(\x_1^n) + \gamma_n(x_n) \}$ needs to be estimated for every $\x_1^{n-1} = (x_1,x_2,\ldots,x_{n-1})$, which is growing exponentially with $n$. Thus from computational perspective the alternating formulation in Theorem \ref{theorem:game} is just as inefficient as the formulation in Theorem \ref{theorem:z}. Nevertheless, this is the building block that enables inference in high-dimensional problems using low dimensional perturbations and max-solvers. Specifically, it provides the means for a new sampling algorithm from the Gibbs distribution and bounds on the log-partition and entropy functions.

\subsection{Ideal Sampling}
\label{sec:unbiased}


Sampling from the Gibbs distribution is inherently tied to estimating the partition function. If we could compute the partition function exactly, then we could sample from the Gibbs distribution sequentially: for dimension $i=1,2, \ldots,n$ sample $x_i$ with probability which is proportional to $\sum_{\x_{i+1}^n} \exp (\pot(\x))$. Unfortunately, directly computing the partition function is \#P-hard. Instead, we construct a family of self-reducible upper bounds which imitate the partition function behavior, namely by bounding the summation over its exponentiations.

\begin{corollary}
\label{corollary:uppers}
Let $\{\gamma_i(x_i) \}_{x_i \in \cX_i, i = 1, 2, \ldots, n}$ be a collection of i.i.d. random variables, each following the Gumbel distribution with zero mean. Set 
	\begin{align}
	\phi_{j}(\x_1^j)= \E_{\gamma} \left[\max_{x_{j+1}^n}  \{ \pot(\x) + \sum_{i=j+1}^n \gamma_i (x_i) \} \right].
	\end{align}
Then for every $j=1,\ldots,n-1$ and every $\x = \x_1^n$ the following inequality holds:
	\begin{align}
	\sum_{x_j} \exp \left( \phi_{j}(\x_1^j) \right) \le  \exp \left( \phi_{j-1}(\x_1^{j-1}) \right).
	\label{eq:upper}
	\end{align}
In particular, for $j=n$ we have $\sum_{x_n} \exp (\pot(\x_1^n))  =  \exp \left( \phi_{n-1}(\x_1^{n-1}) \right)$.  
\end{corollary}

\begin{IEEEproof}
The result is an application of the perturb-max interpretation of the partition function in Theorem \ref{theorem:z}. 
Intuitively, these bounds correspond to moving expectations outside the maximization operations in Theorem \ref{theorem:game}, each move resulting in a different bound. Formally, 
the left hand side can be expanded as
	\begin{align}
	\E_{\gamma_j} \left[ \max_{x_j} \E_{\gamma_{j+1},\ldots,\gamma_n} \left[ \max_{\x_{j+1}^n} \left\{ \pot(\x_1^n) + \sum_{i=j}^n \gamma_i(x_i) \right\} \right] \right],
	\end{align}
while the right hand side is attained by alternating the maximization with respect to $x_j$ with the expectation of $\gamma_{j+1},\ldots,\gamma_n$. The proof then follows by exponentiating both sides.%
\end{IEEEproof}
The above corollary is similar in nature to variational approaches that have been extensively developed to efficiently estimate the partition function in large-scale problems. These are often inner-bound methods where a simpler distribution is optimized as an approximation to the posterior in a KL-divergence sense (e.g., mean field)~\cite{Jordan99}. Variational upper bounds on the other hand are convex, usually derived by replacing the entropy term with a simpler surrogate function and relaxing constraints on sufficient statistics (see, e.g.,~\cite{Wainwright-05upper}).

We use these upper bounds for every dimension $i=1,\ldots,n$ to sample from a probability distribution that follows a summation over exponential functions, with a discrepancy that is described by the upper bound. This is formalized below in Algorithm \ref{alg:unbiased}. Note that $\x = (\x_1^{j-1}, x_j, \x_{j+1}^n)$.

\begin{algorithm}
\caption{Unbiased sampling from Gibbs distribution \label{alg:unbiased}}
\begin{algorithmic}
\Require Potential function $\pot(\x)$, MAP solver
\State Initial step $j = 1$.
\While{$j < n$}
	\State For all $x \in \cX_j$ compute 
		\begin{align}
		\phi_{j}(\x_1^{j-1}, x) = \E_{\gamma}\left[ \max_{\x_{j+1}^n}  \left\{ \pot(\x_1^{j-1}, x, \x_{j+1}^n) + \sum_{i=j+1}^n \gamma_{i}(x_i) \right\} \right].
		\label{eq:ideal_sample}
		\end{align}
	\State Define a distribution on $\cX_j \cup \{r\}$:
		\begin{align}
		p_j(x) &= \frac{ \exp \left( \phi_{j}(\x_1^{j-1}, x)\right) 
			}{ \exp \left( \phi_{j-1}(\x_1^{j-1}) \right) }, \qquad x \in \cX_j \\
		p_j(r) &= 1 - \sum_{x \in \cX_j} p_j(x)
		\end{align}
	\State Sample $x_j$ from $p_j(\cdot)$.
	\If{$x_j = r$}
		\State Set $j = 1$ to restart sampler.
	\Else{ $x_j \in \cX_j$}
		\State Set $j \gets j + 1$.
	\EndIf
\EndWhile \\
\Return $\x = (x_1, x_2, \ldots, x_n)$
\end{algorithmic}
\end{algorithm}

This algorithm is forced to restart the entire sample if it samples the ``reject'' symbol $r$ at any iteration. We say the algorithm accepts if it terminates with an output $\x$. The probability of accepting with particular $\x$ is the product of the probabilities of sampling $x_j$ in round $j$ for $j \in [n]$. Since these upper bounds are self-reducible, i.e., for every dimension $i$ we are using the same quantities that were computed in the previous dimensions $1,2, \ldots,i-1$, we are sampling an accepted configuration proportionally to $\exp(\pot(\x))$, the full Gibbs distribution.  This is summarized in the following theorem.

\begin{theorem}
\label{theorem:unbiased}
Let $p(\x)$ be the Gibbs distribution defined in (\ref{eq:gibbs}) and let $\{\gamma_i(x_i)\}$ be a collection of i.i.d. random variables following the Gumbel distribution with zero mean given in \eqref{eq:gumbelcdf}. Then
	\begin{align*}
	\prob \left( \textrm{Algorithm \ref{alg:unbiased} accepts}  \right)
		=  Z(\theta) \big/ 
			\exp \left( 
			\E_{\gamma} \left[ \max_{\x} \{ \pot(\x) + \sum_{i=1}^n \gamma_i(x_i) \} \right] 
			\right).
	\end{align*}
Moreover, if Algorithm \ref{alg:unbiased} accepts then it produces a configuration $\x = (x_1,\ldots,x_n)$ according to the Gibbs distribution:
	\begin{align*}
	\prob \left( \textrm{Algorithm \ref{alg:unbiased} outputs $\x$} \; \big| \; \textrm{Algorithm \ref{alg:unbiased} accepts} \right) = \frac{\exp(\pot(\x))}{Z(\theta)}.
	\end{align*}
\end{theorem}

\begin{IEEEproof}
Set $\pot_{j}(\x_1^j)$ as in Corollary \ref{corollary:uppers}. The probability of sampling a configuration $\x = (x_1,\ldots,x_n)$ without rejecting is 
	\begin{align*}
	\prod_{j=1}^n 
		\frac{
		\exp \left( \phi_j(\x_1^j) \right) 
		}{
		\exp \left( \phi_{j-1}(\x_1^{j-1}) \right)
		} 
=      \frac{\exp(\pot(\x))
		}{
		\exp \left( \E_{\gamma} 
			\left[ \max_{\x} \left\{ 
				\pot(\x) + \sum_{i=1}^n \gamma_i(x_i) 
				\right\} 
			\right] \right)}.
	\end{align*}
 The probability of sampling without rejecting is thus the sum of this probability over all configurations, i.e., 
 	\begin{align*}
	\prob \left( \textrm{Algorithm \ref{alg:unbiased} accepts}  \right) 
	&=  Z(\theta) \big/ \exp 
	\left( \E_{\gamma} \left[ \max_{\x} \left\{ \pot(\x) + \sum_{i=1}^n \gamma_i(x_i) \right\} \right] \right).
	\end{align*}
Therefore conditioned on acceptance, the output configuration is produced according to the Gibbs distribution. 
 \end{IEEEproof} 

Since acceptance/rejection follows the geometric distribution, the sampling procedure rejects $k$ times with probability $\left(1- \prob\left(\textrm{Algorithm \ref{alg:unbiased} accepts} \right) \right)^k$. The running time of our Gibbs sampler is determined by the average number of rejections $1/\prob(\textrm{Algorithm \ref{alg:unbiased} accepts})$. The exponent of this error event is:
	\begin{align*}
	\log\frac{1}{ \prob\left(\textrm{Algorithm \ref{alg:unbiased} accepts} \right) }
	&= \E_{\gamma} \left[ \max_{\x} \left\{ \pot(\x) + \sum_{i=1}^n \gamma_i(x_i) \right\} \right] - \log Z(\theta).
	\end{align*}
To be able to estimate the number of steps the sampling algorithm requires, we construct an efficiently computable lower bound to the log-partition function that is based on perturb-max values.  

\subsection{Approximate Inference and Lower Bounds to the Partition Function}
\label{sec:approx}

To be able to estimate the number of steps the sampling Algorithm \ref{alg:unbiased} requires, we construct an efficiently computable lower bound to the log-partition function, that is based on perturb-max values. Let $\{M_i : i = 1,2,\ldots, n\}$ be a collection of positive integers. For each $i = 1,2, \ldots, n$ let $\tilde{\x}_i = \{x_{i, k_i} : k_i = 1, 2,\ldots, M_i\}$ be a tuple of $M_i$ elements of $\cX_i$. We define an extended potential function over a configuration space of $\sum_{i=1}^{n} M_i$ variables $\tilde{\x} = (\tilde{\x}_1, \tilde{\x}_2, \ldots, \tilde{\x}_n)$:
	\begin{align}
	\hat \pot (\tilde{\x}) = \frac{1}{\prod_{i=1}^{n} M_i} 
		\sum_{ k_1 =1 }^{ M_1 } 
		\sum_{ k_2 =1 }^{ M_2 }
		\cdots
		\sum_{ k_n =1 }^{ M_n }
		\pot (x_{1,k_1}, x_{2,k_2}, \ldots , x_{n,k_n}).
	\label{eq:ext_pot}
	\end{align}
Now consider a collection of i.i.d. zero-mean Gumbel random variables $\{ \tilde{\gamma}_{i,k_i} (x_{i,k_i}) \}_{i=1,2,\ldots, n, k_i = 1, 2, \ldots, M_i}$ with distribution \eqref{eq:gumbelcdf}. Define the following perturbation for the extended model:
	\begin{align}
	\label{eq:ext_pert}
	\tilde{\gamma}_i(\tilde{\x}_i) = \frac{1}{M_i} \sum_{k_i = 1}^{M_i}  \tilde{\gamma}_{i,k_i} (x_{i,k_i}).
	\end{align}

\begin{corollary}
\label{corollary:p-lower}
Let $\pot(\x)$ be a potential function over $\x = (x_1,\ldots,x_n)$ and $\log Z$ be the log partition function for the corresponding Gibbs distribution. 
Then for any $\epsilon > 0$ we have
	\begin{align}
	\label{eq:ext_bound}
	\prob_{\tilde{\gamma}}\left(\log Z  \ge \max_{ \tilde{\x} } \left\{ \hat \pot (\tilde{\x}) + \sum_{i=1}^n \tilde{\gamma}_i(\tilde{\x}_i) \right\} - \epsilon n \right)
	\ge 1-\sum_{i=1}^n  \frac{ \pi^2 \prod_{j=2}^{i} |\cX_{j-1}| }{6M_i \epsilon^2}.
	\end{align}
\end{corollary}

\begin{IEEEproof} 
The proof consists of three steps:
	\begin{itemize}
	\item developing a measure concentration analysis for Theorem \ref{theorem:z}, which states that a single max-evaluation is enough to lower bound the expected max-value with high probability;
	\item using the self-reducibility of the partition function in Theorem \ref{theorem:game} to show the partition function can be computed by iteratively applying low-dimensional perturbations;
	\item proving that these lower dimensional partition functions can be lower bounded uniformly (i.e., all at once) with a single measure concentration statement.
	\end{itemize}
	
We first provide a measure concentration analysis of Theorem \ref{theorem:z}. Specifically, we estimate the deviation of the random variable $F = \max_{\x \in \cX} \{\pot(\x) + \gum{\x}\}$ from its expected value using Chebyshev's inequality. For this purpose we recall Theorem \ref{theorem:z} which states that $F$ is Gumbel-distributed and therefore its variance is $\pi^2/6$. Chebyshev's inequality then asserts that  
\begin{align}
\label{eq:cheby}
\prob_{\gumset}\left( \left| F - \E_{\gamma}\left[ F \right]  \right|  \ge \epsilon \right) \le \pi^2 /6\epsilon^2.
\end{align}
Since we want this statement to hold with high probability for small epsilon we reduce the variance of the random variable while not changing its expectation by taking a sampled average of i.i.d. perturb-max  values: Let $\mvh(\gamma) = \max_{\x} \{\pot(\x) + \gamma(\x) \}$. Suppose we sample $M$ i.i.d. random variables $\gamma_1, \gamma_2, \ldots, \gamma_M$ with the same distribution as $\gamma$ and generate the  i.i.d.~Gumbel-distributed values $F_j \defeq F(\gamma_j)$. We call $\gamma_1, \gamma_2, \ldots, \gamma_M$ ``copies'' of $\gamma$. Since\footnote{Whenever $\theta$ is clear from the context we use the shorthand $Z$ for $Z(\theta)$.} $\E_\gamma[F(\gamma)] = \log Z$, we can apply Chebyshev's inequality to the $\frac{1}{M} \sum_{i=1}^{M} \mvh_j - \log Z$ to get
	\begin{align}
	\label{eq:chebyM}
	\prob\left( \left| \frac{1}{M} \sum_{i=1}^{M} \mvh_j - \log Z  \right|  \ge \epsilon \right) \le \frac{\pi^2}{6 M\epsilon^2}.
	\end{align}
Using the explicit perturb-max notation and considering only the lower-side of the measure concentration bound, this shows that with probability at least $1- \frac{\pi^2}{6M\epsilon^2}$ we have
\begin{align}
\label{eq:lowerm}
\log Z \ge \frac{1}{M} \sum_{j=1}^M \max_{\x \in \cX} \{\pot(\x) + \gamma_j(\x)\} - \epsilon.
\end{align}
To complete the first step, we wish to compute the summation over $M$-maximum values using a single maximization. For this we form an extended model on $\cX^M$ containing variables $\tilde{\x}_1, \tilde{\x}_2, \ldots, \tilde{\x}_M \in \cX$ and note that
\begin{align}
\label{eq:lower=}
\sum_{j=1}^M \max_{\x \in \cX} \{\pot(\x) + \gamma_j(\x)\} = \max_{\tilde{\x}_1, \tilde{\x}_2, \ldots, \tilde{\x}_M}  \sum_{j=1}^M (\pot(\tilde{\x}_j) + \gamma_j(\tilde{\x}_j) ).
\end{align}

For the remainder we use an argument by induction on $n$, the number of variables. Consider first the case $n = 2$ so that $\pot(\x) = \pot_{1,2}(x_1,x_2)$. 
The self-reducibility as described in Theorem \ref{theorem:z} states that 
\begin{align}
\label{eq:lowerz} 
\log Z = \log \left( \sum_{x_1} \exp \left[ \log \left( \sum_{x_2}   \exp(\pot_{1,2}(x_1,x_2)) \right) \right] \right).
\end{align}
As in the proof of Theorem \ref{theorem:z}, define $\pot_1(x_1) = \log (\sum_{x_2}   \exp(\pot_{1,2}(x_1,x_2))$. Thus we have $\log Z = \log\left( \sum_{x_1} \exp( \pot_1(x_1) ) \right)$, which is a partition function for a single-variable model. 

We wish to uniformly approximate $\pot_1(x_1)$ over all $x_1 \in \mc{X}_1$. Fix $x_1 = a$ for some $a \in \mc{X}_1$ and consider the single-variable model $\pot_{1,2}(a,x_2)$ over $x_2$ which has $\pot_1(a)$ as its log-partition function. Then from Theorem \ref{theorem:z}, we have $\pot_1(a) = \E_{\gamma_2} \left[ \max_{x_2}\{\pot(a,x_2) + \gamma_2(x_2)\} \right]$. Applying Chebyshev's inequality in \eqref{eq:chebyM} to $M_2$ ``copies'' of $\gamma_2$, we get
	\begin{align*}
	\prob\left( \left| \frac{1}{M_2} \sum_{j=1}^{M_2} 
		\max_{x_2}\{\pot(a,x_2) + \gamma_{2,j}(x_2)] \} - \pot_1(a)
		\right| \ge \epsilon 
		\right) 
	\le \frac{\pi^2}{6 M_2 \epsilon^2}.
	\end{align*}
Taking a union bound over $a \in \mc{X}_1$ we have
	\begin{align*}
	\prob\left( \left| \frac{1}{M_2} \sum_{j=1}^{M_2} 
		\max_{x_2}\{\pot(x_1,x_2) + \gamma_{2,j}(x_2)] \} - \pot_1(x_1)
		\right| \le \epsilon \ \ \forall x_1 \in \mc{X}_1
		\right) 
	\le 1 - |\mc{X}_1| \frac{\pi^2}{6 M_2 \epsilon^2}.
	\end{align*}
This implies the following one-sided inequality with probability at least $1 - |\mc{X}_1| \frac{\pi^2}{6 M_2 \epsilon^2}$ uniformly over $x_1 \in \mc{X}_1$:
	\begin{align}
	\label{eq:recurse_pot_1}
	\pot_1(x_1) \ge \frac{1}{M_2} \sum_{j=1}^{M_2} 
		\max_{x_2}\{\pot(x_1,x_2) + \gamma_{2,j}(x_2)] \} - \epsilon.
	\end{align}
Now note that the overall log-partition function for the model $\pot(\x) = \pot_{1,2}(x_1,x_2)$ is a log-partition function for a single variable model with potential $\pot_1(x_1)$, so $\log Z = \log (\sum_{x_1} \exp(\pot_1(x_1)))$. Again using Theorem \ref{theorem:z}, we have $\log Z = \E_{\gamma_1} \left[ \max_{x_1} \{\pot_1(x_1) + \gamma_1(x_1)\}\right]$, so we can apply Chebyshev's inequality to $M_1$ ``copies'' of $\gamma_1$ to get that with probability at least $1 - \frac{\pi^2}{6 M_1 \epsilon^2}$:
	\begin{align}
	\label{eq:recurse_pot_2}
	\log Z \ge \frac{1}{M_1} \sum_{k=1}^{M_1} 
		\max_{x_1}\{\pot_1(x_1) + \gamma_{1,k}(x_1) \} - \epsilon.
	\end{align}
Plugging in \eqref{eq:recurse_pot_1} into \eqref{eq:recurse_pot_2}, we get that with probability at least $1 - \frac{\pi^2}{6 M_1 \epsilon^2} - |\mc{X}_1| \frac{\pi^2}{6 M_2 \epsilon^2}$:
	\[
	\log Z \ge 
		\frac{1}{M_1} \sum_{k=1}^{M_1} 
			\max_{x_1} \left\{ \left(
				\frac{1}{M_2} \sum_{j=1}^{M_2} 
				\max_{x_2} \left\{\pot(x_1,x_2) + \gamma_{2,j}(x_2) \right\} 
				\right)
			+ \gamma_{1,k}(x_1) 
		\right\} 
		- 2\epsilon.
	\]

Now we pull the maximization outside the sum by introducing i.i.d. ``copies'' of the variables again: this time we have $M_1$ copies $\tilde{\x}_1$ and $M_1 M_2$ copies $\tilde{\x}_2$ for  $\tilde{\x}_2$ as in \eqref{eq:lower=}. Now,
	\begin{align*}
	&\frac{1}{M_1} \sum_{k=1}^{M_1} 
			\max_{x_1} \left\{ \left(
				\frac{1}{M_2} \sum_{j=1}^{M_2} 
				\max_{x_2} \left\{\pot(x_1,x_2) + \gamma_{2,j}(x_2) \right\} 
				\right)
			+ \gamma_{1,k}(x_1) 
		\right\} \\
	&= \frac{1}{M_1} \sum_{k=1}^{M_1} 
			\max_{x_1} \left\{ \left(
				\max_{ \tilde{x}_{2,1}, \ldots, \tilde{x}_{2,M_2}} 
				\frac{1}{M_2} \sum_{j=1}^{M_2} 
				\pot(x_1,\tilde{x}_{2,j}) + \gamma_{2,j}(\tilde{x}_{2,j}) 
				\right)
			+ \gamma_{1,k}(x_1) 
		\right\} \\
	&= \max_{ \tilde{x}_{1,1}, \ldots, \tilde{x}_{1,M_1}}  
		\max_{ \tilde{x}_{2,1}, \ldots, \tilde{x}_{2,M_2}}
		 \frac{1}{M_1 M_2} \sum_{k=1}^{M_1}  \sum_{j=1}^{M_2} 
				\pot(\tilde{x}_{1,k},\tilde{x}_{2,j}) 
				+ \gamma_{2,j}(\tilde{x}_{1,k}, \tilde{x}_{2,j}) 
			+ \gamma_{1,k}(\tilde{x}_{1,k}).
	\end{align*}
	
Note that in this bound we have to generate $|\mc{X}_1| |\mc{X}_2|$ variables $\gamma_{2,j}(x_{1,k}, x_{2,j})$, which will become inefficient as we add more variables. We can get an efficiently computable lower bound on this quantity by generating a smaller set of variables: we use the same perturbation realization $\gamma_{2,j}(x_{2,j})$ for every value of $x_{1,k}$. Thus we have the lower bound
	\begin{align*}
	\log Z \ge 
	\max_{ \tilde{\x}_{1}, \tilde{\x}_{2} }  \frac{1}{M_1 M_2} \sum_{k=1}^{M_1}  \sum_{j=1}^{M_2} 
			\left( \pot(x_{1,k},\tilde{x}_{2,j}) + \gamma_{2,j}(\tilde{x}_{2,j}) 
			+ \gamma_{1,k}(\tilde{x}_{1,k}) 
			\right)
	- 2\epsilon
	\end{align*}
with probability at least $1 - \frac{\pi^2}{6 M_1 \epsilon^2} - |\mc{X}_1| \frac{\pi^2}{6 M_2 \epsilon^2}$. Here we have abused notation slightly and used $\tilde{\x}_{1} = \{ \tilde{x}_{1,1}, \tilde{x}_{1,2}, \ldots, \tilde{x}_{1,M_1}\}$ and $\tilde{\x}_2 = \{ \tilde{x}_{2,1}, \tilde{x}_{2,2} \ldots, \tilde{x}_{2,M_2} \}$.

Now suppose the result holds for models on $n-1$ variables and consider the model $\pot(x_1,x_2, \ldots, x_n)$ on $n$ variables. Consider the 2-variable model $\pot(x_1,\x_{2}^{n})$ and define
	\[
	\pot_1(x_1) = \log \left(\sum_{x_{2}^{n}} \exp(\pot(x_1,\x_{2}^{n})) \right).
	\]
From the analysis of the 2-variable case, as in \eqref{eq:recurse_pot_2}, the following lower bound holds with probability at least $1 - \frac{\pi^2}{6 M_1 \epsilon^2}$:
	\[
	\log Z \ge \frac{1}{M_1} \sum_{k_1=1}^{M_1} 
		\max_{x_1}\{\pot_1(x_1) + \gamma_{1,k_1}(x_1) \} - \epsilon.
	\]
Now note that for each value of $x_1$, the function $\pot_1(x_1)$ is a log-partition function on the $n-1$ variables $x_{2}^{n}$. Applying the induction hypothesis to $\pot_1(x_1)$, we have with probability at least 
	\[
	1 - \frac{\pi^2}{6 M_2 \epsilon^2} - |\cX_2| \frac{\pi^2}{6 M_3 \epsilon^2} - |\cX_2| |\cX_3| \frac{\pi^2}{6 M_4 \epsilon^2} - \cdots - \prod_{j=2}^{n-1} |\cX_j|  \frac{\pi^2}{6 M_n \epsilon^2},
	\]
the following lower bound holds:
	\begin{align}
	\pot_1(x_1) \ge \max_{\tilde{\x}_{2}^{n}} \left\{ 
		\hat{\pot}(x_1,\tilde{\x}_2^n)
		+ \sum_{i=2}^{n} \tilde{\gamma}_i(\tilde{\x}_i) \right\}
		- \epsilon (n-1).
	\end{align}
Taking a union bound over all $x_1$, with probability at least 
	\[
	1 - \sum_{i=1}^{n} \left(\prod_{j=2}^{i} |\cX_{j-1}|\right)  \frac{\pi^2}{6 M_n \epsilon^2}
	\]
we have
	\begin{align*}
	\log Z &\ge \frac{1}{M_1} \sum_{k_1=1}^{M_1} 
		\max_{x_1} \left\{
			\max_{\tilde{\x}_{2}^{n}} \left\{ 
				\hat{\pot}(x_1,\tilde{\x}_2^n)
				+ \sum_{i=2}^{n} \tilde{\gamma}_i(\tilde{\x}_i) \right\}
			+ \gamma_{1,k_1}(x_1)
			\right\}
		- \epsilon n \\
	&\ge \max_{\tilde{\x}}
		\hat{\pot}(\tilde{\x}) + \sum_{i=1}^{n} \tilde{\gamma}_i(\tilde{\x}_i) - 
		\epsilon n,
	\end{align*}
as desired.
\end{IEEEproof}

The key to understand the efficiency of this lower bound is in analyzing the structure of the potential functions $\hat \pot (\tilde{\x})$ and $\pot(\x)$. Although $\hat \pot (\tilde{\x})$ seems to consider exponentially many configurations, its order is the same as the original potential function $\pot (\x)$. Particularly, if $\pot (\x)$ is the sum of local and pairwise potential functions (as happens for the Ising model) then $\hat \pot (\tilde{\x})$ is also the sum of  local and pairwise potential functions. Therefore, whenever the original model can be maximized efficiently, e.g., for super-modular functions, the inflated model $\hat \pot (\tilde{\x})$ can also be optimized efficiently. Moreover, while the theory requires $M_i$ to be exponentially large (as a function of $n$), it turns out that in practice $M_i$ may be very small to generate tight bounds (see Section \ref{sec:experiments}). Theoretically tighter bounds can be derived by our measure concentration results in Section \ref{sec:mc} but they do not fully capture the tightness of this lower bound. 

\subsection{Entropy bounds}
\label{sec:entropy}
We now show how to use perturb-max values to bound the entropy of  high-dimensional models. 
Estimating the entropy is an important building block in many machine learning applications. Corollary \ref{corollary:entropy} applies the interpretation of Gibbs distribution as a perturb-max model (see Corollary \ref{corollary:p}) in order to define the entropy of Gibbs distributions using the expected value of the maximal perturbation.  Unfortunately, this procedure requires exponentially many independent perturbations $\gamma(\x)$, for every $\x \in \cX$. 

We again use our low-dimensional perturbations to upper bound the entropy of perturb-max models. We need to extend our definition of perturb-max models as follows. Let $\mc{A}$ be a collection of subsets of $\{ 1,2,\ldots, n\}$ such that $\bigcup_{\alpha \in \mc{A}} = \{1,2,\ldots, n\}$. For each $\alpha \in \mc{A}$ generate a Gumbel perturbation $\gamma_\alpha(\x_\alpha)$ where $\x_\alpha = (x_i)_{i \in \alpha}$. We define the perturb-max models as  
	\begin{align}
	p(\hat \x;\theta) = \prob_{\gamma} \left(\hat{\x} 
		= \argmax_{\x} \left\{ \pot(\x) 
			+ \sum_{\alpha \in \mc{A}} \gamma_\alpha(\x_\alpha) 		\right\} \right).  \label{eq:model} 
	\end{align}
Our upper bound uses  the duality between entropy and the log-partition function~\cite{Wainwright08} and then upper bounds the log-partition function with perturb-max operations.

Upper bounds for the log-partition function using random perturbations can be derived from the refined upper bounds in Corollary \ref{corollary:uppers}. However, it is simpler to provide upper bounds that rely on Theorem \ref{theorem:game}. These bounds correspond to moving expectations outside the maximization operations. 

\begin{lemma}
\label{lemma:upper}
Let $\pot(\x)$ be a potential function over $\x = (x_1,\ldots,x_n)$, and $\{\gamma_i(x_i)\}_{x_i\in \cX_i, i=1,\ldots,n}$ be a collection of independent and identically distributed (i.i.d.) random variables following the Gumbel distribution. Then 
	\begin{align}
	\log Z(\theta) \le \E_{\gamma} \left[\max_{\x = (x_1,x_2, \ldots,x_n)} \left\{\pot(\x) + \sum_{i=1}^n \gamma_i(x_i) \right\} \right]. \label{basic:upper}
	\end{align}
\end{lemma}
\begin{IEEEproof}
The lemma follows from Theorem \ref{theorem:game} that represents \eqref{eq:Emax} as the log-partition as a sequence of alternating expectations and maximizations, namely  
	\begin{align}
	\log Z(\theta) = \E_{\gamma_1} \max_{x_1} \cdots \E_{\gamma_n} \max_{x_n}  \left\{ \pot(\x) + \sum_{i=1}^n \gamma_i(x_i) \right\}. 
	\end{align}
The upper bound is attained from the right hand side of the above equation by Jensen's inequality (or equivalently, by moving all the expectations in front of the maximizations, yielding the following:
	\begin{align}
	\E_{\gamma_1} \max_{x_1} \cdots \E_{\gamma_n} \max_{x_n}  \left\{ \pot(\x) + \sum_{i=1}^n \gamma_i(x_i) \right\} \le \E_{\gamma_1} \cdots \E_{\gamma_n} \max_{x_1}  \cdots \max_{x_n}  \left\{ \pot(\x) + \sum_{i=1}^n \gamma_i(x_i) \right\}.
	\end{align}
\end{IEEEproof}

In this case the bound is an average of MAP values corresponding to models with only single node perturbations $\gamma_i(x_i)$, for every $i=1,\ldots,n$ and $x_i\in \cX_i$. If the maximization over $\pot(\x)$ is feasible (e.g., due to supermodularity), it will typically be feasible after such perturbations as well.  We generalize this basic result further below.

\begin{corollary}
\label{corollary:upper}
Consider a family of subsets $\alpha \in \mc{A}$ such that $\bigcup_{\alpha\in \mc{A}} \alpha = \{1,\ldots,n\}$, and let $\x_\alpha = \{x_i : i \in \alpha \}$. Assume that the random variables $\gamma_\alpha(\x_\alpha)$ are i.i.d.  according to the Gumbel distribution, for every $\alpha,\x_\alpha$. Then 
	\begin{align*}
	\log Z(\theta) \le \E_{\gamma} \left[\max_{\x} \left\{\pot(\x) + \sum_{\alpha\in \mc{A}} \gamma_\alpha(\x_\alpha) \right\} \right].
	\end{align*} 
\end{corollary}

\begin{IEEEproof}
If the subsets $\alpha$ are disjoint the upper bound is an application of Lemma \ref{lemma:upper} as follows: we consider the potential function $\pot(\x)$ over the disjoint subsets of variables $\x = (\x_\alpha)_{\alpha \in {\cal A}}$ as well as the i.i.d. Gumbel random variables $\gamma_\alpha(\x_\alpha)$. Applying Lemma \ref{lemma:upper} yields the following upper bound: 
	\begin{align}
	\log Z(\theta) \le \E_{\gamma} \left[  \max_{\x = (\x_\alpha)_{\alpha \in {\cal A}}} \left\{\pot(\x) + \sum_{\alpha \in {\cal A}} \gamma_\alpha(\x_\alpha) \right\} \right].
	\end{align}
In the general case, $\alpha, \beta \in \mc{A}$ may overlap. To follow the same argument, we lift the $n$-dimensional assignment $\x = (x_1,x_2,\ldots,x_n)$ to an higher-dimensional assignment $a(\x) = (\x_\alpha)_{\alpha \in \mc{A}}$ which creates an independent perturbation for each $\alpha \in \mc{A}$. To complete the proof, we also construct a potential function $\pot'(\x')$ such that 
	\begin{align}
	\pot'(\x') = \begin{cases}
	\pot(\x) & \textrm{if $a(\x)=\x'$} \\
	 -\infty & \textrm{otherwise.}
\end{cases}
	\end{align}
Thus, $\log Z(\pot) = \log Z(\pot') =\sum_{\x'} \exp(\pot'(\x'))$ since inconsistent assignments (i.e., $\x'$ such that $a(\x) \ne \x'$ for any $\x$) receive zero weight. Moreover,
	\begin{align*}
	\max_{\x'} \left\{ \pot'(\x') + \sum_{\alpha \in \mc{A}} \gamma_\alpha(\x'_\alpha) \right\}
	=
	\max_{\x} \left\{ \pot(\x) + \sum_{\alpha \in \mc{A}} \gamma_\alpha(\x_\alpha) \right\}
	\end{align*}
for each realization of the perturbation. This equality holds after expectation over $\gamma$ as well. Now, given that the perturbations are independent for each lifted coordinate, the basic result in (\ref{basic:upper}) guarantees that 
	\begin{align*}
	\log Z(\pot')
	\le 
	\E_{\gamma} \left[\max_{\x'}  \left\{\pot'(\x') + \sum_{\alpha \in \mc{A}} \gamma_\alpha(\x'_\alpha) \right\} \right],
	\end{align*}
from which the result follows since $\log Z(\pot) = \log Z(\pot')$
\end{IEEEproof}

Establishing bounds on the log-partition function allows us to derive bounds on the entropy. For this we use the conjugate duality between the (negative) entropy and the log-partition function~\cite{Wainwright08}. The entropy bound then follows from the log-partition bound.
   
\begin{theorem}
\label{theorem:bound}
Let $p(\x; \theta)$ be a perturb-max probability distribution in  (\ref{eq:model}) and $\mc{A}$ be a collection of subsets of $\{1,2,\ldots, n\}$. Let $\x^{\gamma}$ be the optimal perturb-max assignment using low dimensional perturbations:
	\begin{align}
	\x^{\gamma} = \argmax_{\x} \left\{ \pot(\x) + \sum_{\alpha \in \mc{A}} \gamma_\alpha(\x_\alpha) \right\}.
	\label{eq:opt_perturb_max}
	\end{align} 
Then under the conditions of Corollary \ref{corollary:upper}, we have the following upper bound:
	\begin{align*}
H(p) \le \E_\gamma \left[ \sum_{\alpha \in \mc{A}} \gamma_\alpha(\x_{\alpha}^{\gamma}) \right].
	\end{align*}
\end{theorem}

\begin{IEEEproof}
We use the characterization of the log-partition function as the conjugate dual of the (negative) entropy function~\cite{Wainwright08}: 
	\begin{align*}
	H(p) = \min_{\hat{\pot}} \left\{ \log Z(\hat{\pot}) - \sum_{\x} p(\x; \pot) \hat{\pot}(\x) \right\}.
	\end{align*}
The minimum is over all potential functions on $\mc{X}$.
For a fixed score function $\hat{\pot}(\x)$, let $W(\hat{\pot})$ be the expected value of the low-dimensional perturbation:
	\begin{align*}
	W(\hat{\pot}) = \E_\gamma \left[ \max_{\x} \left\{\hat{\pot}(\x) + \sum_{\alpha \in \mc{A}} \gamma_\alpha(\x_\alpha) \right\} \right].
	\end{align*}
Corollary \ref{corollary:upper} asserts that $\log Z(\hat{\pot}) \le W(\hat{\pot})$. Thus we can upper bound $H(p)$ by replacing $\log Z(\hat{\pot})$ with $W(\hat{\pot})$ in the duality relation:
	\begin{align*}
	H(p) \le \min_{\hat{\pot}} \left\{ W(\hat{\pot}) - \sum_{\x} p(\x; \pot) \hat{\pot}(\x) \right\}.
	\end{align*} 
The infimum of the right hand side is attained whenever the gradient vanishes, i.e., whenever $\nabla W(\hat{\pot}) = p(\x; \pot)$.
To compute $\nabla W(\hat{\pot})$ we differentiate under the integral sign:
	\begin{align*}
	\nabla W(\hat{\pot}) = \E_\gamma \left[ \nabla \max_{\x} \left\{\hat{\pot}(\x) + \sum_{\alpha \in \mc{A}} \gamma_\alpha(\x_\alpha) \right\} \right]. 
	\end{align*}
Since the (sub)gradient of the maximum-function is the indicator function, we deduce that $\nabla W(\hat{\pot})$ is the expected value of the events of $\x^{\gamma}$. Consequently, $\nabla W(\hat{\pot})$ is the vector of the probabilities of all these events, namely, the probability distribution $p(\x; \hat \pot)$.
Since the derivatives of $W(\hat{\pot})$ are perturb-max models, and so is $p(\x; \pot)$, then the the infimum is attained for $\hat{\pot} = \pot$. Therefore, recalling that $\x^{\gamma}$ has distribution $p(\x;\pot)$ in \eqref{eq:model}:
	\begin{align*}
	\min_{\hat{\pot}} \left\{ W(\hat{\pot}) 
		- \sum_{\x} p(\x; \pot) \pot(\x) \right\} 
	&= W(\pot) - \sum_{\x} p(\x; \pot)  \pot(\x). \\
	&= \E_\gamma \left[ \max_{\x} \left\{ \pot(\x) + \sum_{\alpha \in \mc{A}} \gamma_\alpha(\x_\alpha) \right\} \right]
		- \E_\gamma \left[ \pot(\x^{\gamma}) \right] \\
	&= \E_\gamma \left[ 
		\hat{\pot}(\x^{\gamma}) + \sum_{\alpha \in \mc{A}} \gamma_\alpha(\x_\alpha^{\gamma}) 
		\right]
		- \E_\gamma \left[ \pot(\x^{\gamma}) \right] \\
	&= \E_\gamma \left[ 
		\sum_{\alpha \in \mc{A}} \gamma_\alpha(\x_\alpha^{\gamma}) 
		\right],
	\end{align*}
from which the result follows.
\end{IEEEproof}

This entropy bound motivates the use of perturb-max posterior models. These models are appealing as they are uniquely built around prediction and as such they inherently have an efficient unbiased sampler. The computation of this entropy bound relies on MAP solvers. Thus, computing these bounds is significantly faster than computing the entropy itself, whose computational complexity is generally exponential in $n$.   

Using the linearity of expectation we may alternate summation and expectation. For simplicity, assume only local perturbations, i.e., $\gamma_i(x_i)$ for every dimension $i=1,\ldots,n$. Then the preceding theorem bounds the entropy by summing the expected change of MAP perturbations $H(p) \le \sum_i \E_\gamma  [\gamma_i(x^{\gamma}_i) ]$. This bound resembles to the independence bound for the entropy $H(p) \le \sum_i H(p_i)$, where $p_i(x_i) = \sum_{\x \setminus x_i} p(\x)$ are the marginal probabilities~\cite{Cover12}. The independence bound is tight whenever the joint probability $p(\x)$ is composed of independent systems, i.e., $p(\x) = \prod_i p_i(x_i)$. In the following we show that the same holds for perturbation bounds.

\begin{corollary}
Consider the setting in Theorem \ref{theorem:bound} with $\x^{\gamma}$ given by \eqref{eq:opt_perturb_max} and the independent probability distribution $p(\x) = \prod_i p_i(x_i)$. Let $\{\gamma_i(x_i) \}_{x_i \in \cX_i, i = 1, 2, \ldots, n}$ be a collection of i.i.d. random variables, each following the Gumbel distribution with zero mean. Then there exists $\pot(\x)$ for which  
\[
H(p) = \E_\gamma \left[ \sum_{i=1}^n \gamma_i(x^{\gamma}_i) \right], \nonumber 
\]
where 
	\begin{align*}
	\x^{\gamma} =  \argmax_{\x} \left\{ \pot(\x) + \sum_{i=1}^n \gamma_i(x_i) \right\}.
	\end{align*}
\end{corollary}   
\begin{IEEEproof} 
Since the system is independent, $H(p) = \sum_i H(p_i)$. We first show that there exists $\pot_i(x_i)$ in each dimension for which $H(p_i) = \E_{\gamma_i} \left[\gamma_i(x^{\gamma_i}_i) \right]$ and then complete the proof by constructing $\pot(\x)$. 

Set $\pot_i(x_i) = \log p_i(x_i)$. Since $\{\gamma_i(x_i) \}_{x_i \in \cX_i}$ are independent, we may apply Corollary \ref{corollary:entropy} to the $i-$th dimension and obtain $H(p_i) = \E_\gamma \left[\gamma_i(x^{\gamma_i}_i) \right]$, where $x^{\gamma_i} = \argmax_{\x_i} \left\{ \pot_i(\x_i) + \gamma_i(x_i) \right\}$. 

To complete the proof, we set $\pot(\x) = \sum_i \pot_i(x_i)$. Since the system is independent, there holds $\x^{\gamma}_i =  \x^{\gamma_i}_i  \stackrel{def}{=} \argmax_{\x_i} \left\{ \pot_i(\x_i) + \gamma_i(x_i) \right\} $. Therefore, $H(p) = \sum_i H(p_i) = \sum_i  \E_{\gamma_i} \left[\gamma_i(x^{\gamma_i}_i) \right] =  \sum_i  \E_{\gamma} \left[\gamma_i(x^{\gamma}_i) \right] =  \E_{\gamma} \left[\sum_i  \gamma_i(x^{\gamma}_i) \right]$.  
\end{IEEEproof}

There are two special cases for independent systems. First, the zero-one probability model, for which $p(\x) = 0$ except for a single configuration $p(\hat{\x})=1$. The entropy of such a probability distribution is $0$ since the distribution is deterministic. In this case, the perturb-max entropy bound assigns $\x^{\gamma} = \hat{\x}$ for all random functions $\gamma = (\gamma_i(x_i))_{i,x_i}$. Since these random variables have zero mean, it follows that $\E_\gamma \left[ \sum_i \gamma_i(\hat{x}_i) \right] = 0$. Another important case is for the uniform distribution with $p(\x) = 1/|\cX|$ for every $\x \in \cX$. The entropy of such a probability distribution is $\log |\cX|$, as it has maximal uncertainty. Since our entropy bounds equal the entropy for minimal uncertainty and maximal uncertainty cases, this suggests that the perturb-max bound can be used as an alternative uncertainty measure.
\begin{corollary}
Consider the setting of Theorem \ref{theorem:bound} with $\x^{\gamma}$ given by \eqref{eq:opt_perturb_max}.  Define the function $U(p)$ by
\[
U(p) = \E_\gamma \left[ \sum_{\alpha \in \mc{A}} \gamma_\alpha(\x_{\alpha}^{\gamma}) \right].
\]
Then $U(p)$ is non-negative and attains its minimal value for the deterministic distributions and its maximal value for the uniform distribution. 
\end{corollary}   

\begin{IEEEproof}
 As argued above, $U(p)$ is $0$ for deterministic $p$. Non-negativity follows from the requirement that the perturbation are zero-mean random variables: since $ \sum_\alpha \gamma_\alpha (\x^\gamma_\alpha) \ge \sum_\alpha \gamma_\alpha (\x_\alpha)$ for $x$, then $U(p) =  \E_\gamma \sum_\alpha \gamma_\alpha (\x^\gamma_\alpha) \ge  \E_\gamma \sum_\alpha \gamma_\alpha (\x_\alpha) = 0$. Lastly, we must show that the uniform distribution $p_{\mathsf{uni}}$ maximizes $U(\cdot)$, namely $U(p_{\mathsf{uni}}) \ge U(\cdot)$. The potential function for the uniform distribution is constant for all $\x \in \cX$. This means that $U( p_{\mathsf{uni}} ) =  \E_\gamma \max_x \sum_\alpha \gamma_\alpha (\x_\alpha)$. On the other hand $U(\cdot)$ correspond to a potential function $\pot(\x)$ and its corresponding $\x^\gamma$. Furthermore, we have $\max_{\x} \sum_\alpha \gamma_\alpha (\x_\alpha) \ge \sum_\alpha \gamma_\alpha (\x^\gamma_\alpha)$, and taking expectations on both sides shows $U( p_{\mathsf{uni}} ) = \E_\gamma \max_x \sum_\alpha \gamma_\alpha (\x_\alpha) \ge \E_\gamma \sum_\alpha \gamma_\alpha (\x^\gamma_\alpha)$ for any other $\theta(\cdot)$ and its corresponding $\x^\gamma$. 
\end{IEEEproof}

The preceding result implies we can use $U(p)$ as a surrogate uncertainty measure instead of the entropy. Using efficiently computable uncertainty measures allows us to extend the applications of perturb-max models to Bayesian active learning~\cite{Maji14}. The advantage of using the perturb-max uncertainty measure over the entropy function is that it does not require MCMC sampling procedures. Therefore, our approach fits well with contemporary techniques for using high-dimensional models that are popular in machine learning applications such as computer vision. Moreover, our perturb-max uncertainty measure is an upper bound on the entropy, so minimizing the upper bound can be a reasonable heuristic approach to reducing entropy.

\section{Measure concentration for log-concave perturbations \label{sec:mc}}

High dimensional inference with random perturbations relies on expected values of MAP predictions. In Section \ref{sec:unbiased} we presented a Gibbs distribution sampler that involves calculating the expected value of randomized max-solvers $F(\gamma) = \max_{\x} \{\theta(\x) + \sum_\alpha \gamma_\alpha(\x_\alpha) \}$. In Section \ref{sec:entropy} the expected value of the perturbation themselves gave an upper bound on the entropy of the perturb-max model $F(\gamma) = \sum_\alpha \gamma_\alpha(\x^{\gamma}_\alpha)$. Practical application of this theory requires estimating these expectations; the simplest way to do this is by taking a sample average. We therefore turn to bounding the number of samples by proving concentration of measure results for our random perturbation framework.
The key technical challenge comes from the fact that the perturbations $\gamma_\alpha(\x_\alpha)$ are Gumbel random variables, which have support on the entire real line.  Thus, many standard approaches for bounded random variables, such as McDiarmid's inequality, do not apply. 

Nevertheless, the Gumbel distribution decays exponentially, thus one can expect that the distance between the perturbed MAP prediction and its expected value to decay exponentially as well.  
Our measure concentration bounds (in Section \ref{sec:bounds}) show this; we bound the deviation of a general function $\mvh(\gamma)$ of Gumbel variables via its moment generating function 
	\begin{align}
	\Lambda_{\mvh} (\lambda) \defeq \expe{ \exp(\lambda \mvh) }.
	\label{eq:mgf}
	\end{align}
For notational convenience we omit the subscript when the function we consider is clear from its context. The exponential decay follows from the Markov inequality: $\prob\left( \mvh(\gamma) \ge r \right) \le \Lambda(\lambda) / \exp( \lambda r)$ for any $\lambda > 0$.

We derive bounds on the moment generating function $\Lambda(\lambda)$ (in Section \ref{sec:bounds}) by looking at the expansion (i.e., gradient) of $F(\gamma)$. Since the max-value changes at most linearly with its perturbations, its expansion is bounded and so is $\Lambda(\lambda)$. Such bounds have gained popularity in the context of isoperimetric inequalities, and series of results have established measure concentration bounds for general families of distributions, including log-concave distributions~\cite{BrascampL76, Aida94, BobkovL:97exp, Ledoux:01concentration, Bakry13, Nguyen14}. The family of log-concave distribution includes the Gaussian, Laplace, logistic and Gumbel distributions,  among many others. 
A one dimensional density function $q(t)$ is said to be log-concave if $q(t) = \exp(-Q(t))$ and $Q(t)$ is a convex function: log-concave distributions have log-concave densities.
These probability density functions decay exponentially\footnote{One may note that for the Gaussian distribution $Q'(0) = 0$ and that for Laplace distribution $Q'(0)$ is undefined. Thus to demonstrate the exponential decay one may verify that $Q(t) \ge Q(c) + t Q'(c)$ for any $c$ thus $q(t) \le q(c)\exp(-t Q'(c))$.} with $t$. To see that we recall that for any convex function $Q(t) \ge Q(0) + t Q'(0)$ for any $t$. By exponentiating and rearranging we can see $q(t) \le q(0)\exp(-t Q'(0))$.

\subsection{A Poincar\'{e} inequality for log-concave distributions}
\label{sec:poincare}

A Poincar\'{e} inequality bounds the variance of a random variable by its expected expansion, i.e., the norm of its gradient. These results are general and apply to any (almost everywhere) smooth real-valued functions $f(t)$ for $t \in \R^m$. The variance of a random variable (or a function) is its square distance from its expectation, according to the measure $\mu$:
	\begin{align}
	\Var_{\mu}(f) \defeq \int f^2(t) d \mu(t) - \left(\int f(t) d \mu(t) \right)^2. 
	\end{align}
A Poincar\'{e} inequality is a bound of the form 
	\begin{align}
	\Var_{\mu}(f) \le C \int \|\nabla f(t)\|^2 d \mu(t).
	\end{align}
If this inequality holds for any function $f(t)$ we say that the measure $\mu$ satisfies the Poincar\'{e} inequality with a constant $C$. The optimal constant $C$ is called the Poincar\'{e} constant. To establish a Poincar\'{e} inequality it suffices to derive an inequality for a one-dimensional function and extend it to the multivariate case by tensorization~\cite[Proposition 5.6]{Ledoux:01concentration}. 

Restricting to one-dimensional functions, $\Var_{\mu}(f) \le \int_{-\infty}^\infty f(t)^2 q(t) d t $ and the one-dimensional Poincar\'{e} inequality takes the form: 
	\begin{align}
	\int_{-\infty}^\infty f(t)^2 q(t) d t \le C \int_{-\infty}^\infty f'(t)^2 q(t) dt.
	\end{align} 
We provide an elementary proof that is based on the seminal work of Brascamp and Lieb~\cite{BrascampL76}.
We begin by considering a simpler setting, where $Q'(t) \ne 0$. This setting demonstrates the core idea of our general proof while avoiding technical complications. 
\begin{lemma}
\label{lemma:poincare}
Let $\mu$ be a log-concave measure with density $\lcdens(t)=\exp(-\cvxf(t))$, where $\cvxf: \R \rightarrow \R$ is a convex function, twice continuously differentiable almost everywhere, satisfying $Q'(t) \ne 0 \ \forall t$, $\lim_{t\rightarrow +\infty} Q'(t)\geq0$, and $\lim_{t\rightarrow -\infty} Q'(t)\leq0$. 
Let $\h : \R \rightarrow \R$ be a continuous function in $L^2(\mu)$, differentiable almost everywhere and with $\h' \in L^2(\mu)$.
Then for any $\eta \in \left[\displaystyle \min_{t \in \R} -\frac{\cvxf''(t)}{(\cvxf'(t))^2}, 1\right]$, we have 
\begin{align*}
	\Var_{\mu}(\h) \leq \frac{1}{1-\eta}\int_{\R} \frac{(\h'(t))^2}{\cvxf''(t)+ \eta (\cvxf'(t))^2} \lcdens(t) dt.
\end{align*}
\end{lemma}
\begin{IEEEproof}
The variance of $f(t)$ is upper bounded by its second moment, so it suffices to prove that $\int_{-\infty}^\infty f^2(t) q(t) d t \le  \int_{-\infty}^\infty C(t) f'(t)^2 q(t) dt$. We define the function $\psi(t)=\left( h(t)g(t) \right)'$ with $h(t) = f^2(t)$ and $g(t) = q(t)/Q'(t)$. Its integral is nonnegative since $\int \psi(t) dt = \int \left( h(t)g(t) \right)' dt =  \lim_{t \rightarrow \infty} f^2(t) q(t)/Q'(t) - \lim_{t \rightarrow -\infty} f^2(t) q(t)/Q'(t) \geq 0$ by the assumptions on the limits of $Q'(t)$.


The main challenge of the proof is to show the following bound on the function $ \psi(t)$: 
\begin{equation}
\label{eq:poincare-proof}
\psi(t)=\left(f^2(t) \cdot \frac{q(t)}{Q'(t)} \right)' \le -q(t) f^2(t) + q(t) \eta f^2(t) + q(t) \frac{f'^2(t)}{Q''(t) + \eta Q'^2(t)}.
\end{equation}
Assuming that (\ref{eq:poincare-proof}) holds, the proof then follows by taking an integral over both sides, while noticing that the left hand side is nonnegative. 
To prove the inequality in (\ref{eq:poincare-proof}) we first note that by differentiating the function $\psi(t)$, we get
\begin{align}
	\left[f^2(t) \cdot \frac{q(t)}{Q'(t)}\right]' = -q(t) f^2(t) - q(t) f^2(t) \frac{Q''(t)}{Q'^2(t)} + 2f(t) f'(t) \frac{q(t)}{Q'(t)}.
\end{align}
Using the inequality $2 ab \le c a^2 + b^2/c$ for any $c \ge 0$ we derive the bound 
\begin{align}
	2 \frac{f'(t)}{Q'(t)}  f(t) \le  \left( c(t) f^2(t) +  \frac{f'^2(t)}{c(t) Q'^2(t)} \right).
\end{align}
Finally, we set $c(t) = \frac{Q''(t)}{Q'^2(t)} + \eta$ to satisfy $c(t) \ge 0$ and get the inequality in (\ref{eq:poincare-proof}). 
\end{IEEEproof}
The above proof relies on the fact that $Q'(t) \ne 0$ to ensure that the function $\int \left( h(t)g(t) \right)' dt$ is nonnegative.
This holds, for example, for the Laplace distribution $q(t) = \exp(-|t|)/2$ where $Q'(t) \in \{-1,1\}$.
In particular, the Poincar\'{e} inequality for the Laplace distribution given by Ledoux~\cite{Ledoux:01concentration} follows from our result by setting $\eta = 1/2$.

Unfortunately, the condition $Q'(t) \ne 0$ does not hold for all log-concave measures. For example, the Gaussian distribution has $Q'(0) = 0$ and for the Gumbel distribution $Q'(-c) = 0$. The proof above fails in these cases since the function $\left( h(t)g(t) \right)'$ tends to infinity around the point of singularity, namely $a$ for which $Q'(a)=0$. To overcome the singularity $Q'(a) = 0$ we modify the function $f(t)$ such that it vanishes at $a$ in such a way that the numerator (namely $f(a)$) approaches zero faster than the denominator (namely $Q'(a)$).   

\begin{theorem}
\label{theorem:poincare}
Let $\mu$ be a log-concave measure with density $\lcdens(t)=\exp(-\cvxf(t))$, where $\cvxf: \R \rightarrow \R$ is a convex function that has a unique minimum in the point $t=a$. Also, assume $Q(t)$ is twice continuously differentiable excepts possibly at $t=a$, $\lim_{t \rightarrow a^\pm} \cvxf'(t)\neq0$ or $\lim_{t\rightarrow a^\pm} \cvxf''(t)\neq0$, $\lim_{t\rightarrow +\infty} Q'(t)\geq0$, and $\lim_{t\rightarrow -\infty} Q'(t)\leq0$.
Let $\h : \R \rightarrow \R$ be a continuous function in $L^2(\mu)$, differentiable almost everywhere and with $\h' \in L^2(\mu)$.
Then for any $\eta \in \left[\min_{t \in \R\setminus \{a\}} - \frac{\cvxf''(t)}{(\cvxf'(t))^2}, 1\right]$, we have 
\begin{equation*}
  \Var_{\mu}(\h) \leq \frac{1}{1-\eta}\int_{\R} \frac{(\h'(t))^2}{\cvxf''(t)+ \eta (\cvxf'(t))^2} \lcdens(t) dt.
\end{equation*}
\end{theorem}
\begin{IEEEproof}
To ensure that $f(a) = 0$ we use the a different bound on the variance. Specifically, $\Var (f) \le \int_{-\infty}^\infty (f(t) - K)^2 q(t) d t$ for any $K$. Thus we set $K=f(a)$ and follow the proof of Lemma~\ref{lemma:poincare} with $\hat f(t) = f(t) - f(a)$. Since $f'(t) =\hat f'(t)$ an inequality of the form 
\begin{align}
	\int_{-\infty}^\infty \hat f(t)^2 q(t) d t \le \frac{1}{1-\eta}\int_{\R} \frac{(\hat \h'(t))^2}{\cvxf''(t)+ \eta (\cvxf'(t))^2} \lcdens(t) dt
\end{align}
provides the desired Poincar\'{e} inequality. 

To complete the proof, we show that $\int \left( h(t)g(t) \right)' dt$ is nonnegative while $h(t) = \hat f^2(t)$ and $g(t) = q(t)/Q'(t)$. We do so by dividing it into two components with respect to the point $a$: 
\begin{align}
	\int_{-\infty}^\infty \left( \hat f^2(t) \cdot \frac{q(t)}{Q'(t)} \right)' dt = \int_{-\infty}^a \left( \hat f^2(t) \cdot \frac{q(t)}{Q'(t)} \right)' dt + \int_a^\infty \left( \hat f^2(t) \cdot \frac{q(t)}{Q'(t)} \right)' dt.
\end{align}
Lastly,  
\begin{align}
	\int_{-\infty}^a \left( \hat f^2(t) \cdot \frac{q(t)}{Q'(t)} \right)' dt =  \lim_{t \rightarrow a} \hat f^2(t) q(t)/Q'(t) - \lim_{t \rightarrow -\infty} \hat f^2(t) q(t)/Q'(t).
\end{align} 
As in the proof of Lemma~\ref{lemma:poincare}, we have that $\lim_{t \rightarrow -\infty} \hat f^2(t) q(t)/Q'(t) \geq 0$.
The treatment of the term $\lim_{t \rightarrow a} \hat f^2(t) q(t)/Q'(t)$ is slightly more involved since both $Q'(a) = 0$ and $\hat f(a) = 0$. Thus to evaluate the limit we use L'H\^{o}pital's rule and differentiate both the numerator and denominator to obtain $2 \hat f(t) \hat f'(a) q(a)/Q''(a) = 0$ by the assumption that either $Q'(a)=0$ or $Q''(a)=0$ but not both. The same argument follows for the integral over the interval $[a,\infty]$.
\end{IEEEproof}

Brascamp and Lieb~\cite{BrascampL76} proved a Poincar\'{e} inequality for strongly log-concave measures, where $Q''(t) \ge c$. Their result may be obtained by our derivation when $\eta = 0$. Their result was later extended by Bobkov~\cite{Bobkov09} and more recently by Nguyen~\cite{Nguyen14} to log-concave measures and to multivariate functions while restricting their content to the interval $\eta \in [1/2,1]$. In the one-dimensional setting, our bound is tighter, i.e., for $\eta \in \left[\min_{t \in \R\setminus \{a\}} - \frac{\cvxf''(t)}{(\cvxf'(t))^2}, 1\right]$. 

\subsection{Bounds for the Gumbel distribution}

For the Gumbel distribution we get the following bound. 
\begin{corollary}[Poincar\'{e} inequality for the Gumbel distribution]
\label{cor:cor1}
Let $\mu$ be the measure corresponding to the Gumbel distribution. Let $\h: \R^m \rightarrow \R$ be a multivariate function that satisfies the conditions in Lemma~\ref{lemma:poincare} for each dimension. Then
	\begin{align}
	\Var_{\mu}(\h) \leq 4 \int_{\R} \|\nabla \h(t)\|^2 d\mu(t).
	\label{eq:gumbel:poincare}
	\end{align}
\end{corollary}
\begin{IEEEproof}
First, we derive the Poincar\'{e} constant for a one dimensional Gumbel distribution. Then we derive the multivariate bound by tensorization. Following Theorem \ref{theorem:poincare} it suffices to show that there exists $\eta$ such that $\frac{1}{(1-\eta) (\cvxf''(t)+ \eta (\cvxf'(t))^2)} \le 4$ for any $t$. 

For the Gumbel distribution,
	\begin{align}
	\cvxf(t) &= t + c + \exp( - (t + c) ) \\
	\cvxf''(t) + \eta( \cvxf'(t) )^2 &= e^{-(t+c)} + \eta ( 1 - e^{-(t+c)} )^2. \label{eq:poin:gumb1}
	\end{align}
Simple calculus shows that $t^*$ minimizing \eqref{eq:poin:gumb1} is given by
	\begin{align}
	0 &= -(t^*+c) e^{-(t^*+c)} - 2 \eta (t^*+c) ( 1 - e^{-(t^*+c)} ) e^{-(t^*+c)}
	\end{align}
or $e^{-(t^*+c)} = 1 - \frac{1}{2 \eta}$.
The lower bound is then $\cvxf''(t) + \eta( \cvxf'(t) )^2 \ge \frac{4 \eta - 1}{4 \eta}$ whenever $1 - \frac{1}{2 \eta}$ is positive, or equivalently whenever $\eta > \frac{1}{2}$.  For $\eta \le \frac{1}{2}$, we note that $\cvxf''(t) + \eta( \cvxf'(t) )^2 = \eta + (1 - 2 \eta) e^{-(t+c)} + \eta e^{-2(t+c)} \ge \eta$.

Combining these two cases, the Poincar\'{e} constant is at most $\min\left\{ \frac{4 \eta}{(4 \eta - 1)(1 - \eta)}, \frac{1}{\eta (1 - \eta)} \right\} = 4$ at $\eta = \frac{1}{2}$. By applying Theorem \ref{lemma:poincare} we obtain the one-dimensional Poincar\'{e} inequality.

Finally, for $f:\mathbb{R}^m \rightarrow \mathbb{R}$ we denote by $\Var_i(f)$ the variance of $i$-th variable while fixing the rest of the $m-1$ variables in $f(t_1,t_2,\ldots,t_m)$. The one dimensional Poincar\'{e} inequality implies that 
	\begin{align}
	\Var_i(f) \le 4  \int_{\R} |\partial \h(t) / \partial t_i|^2 d\mu. 
	\end{align}
The proof then follows by a tensorization argument given by Ledoux~\cite[Proposition 5.6]{Ledoux:01concentration}, which shows that
	\begin{align}
	\Var_{\mu} (f) \le \sum_{i=1}^m \mathbb{E}_{t \setminus t_i} \left[ \Var_i(f) \right].
	\end{align}
\end{IEEEproof}

Although the Poincar\'{e} inequality establishes a bound on the variance of a random variable, it may also be used to bound the moment generating function $\Lambda_f(\lambda) = \int \exp(\lambda f(t)) d \mu(t)$~\cite{BobkovL:97exp,Aida94,Bakry13}. For completeness we provide a proof specifically for the Gumbel distribution.
   
\begin{corollary}[MGF bound for the Gumbel distribution]
\label{cor:p-mgf}
Let $\mu$ be the measure corresponding to the Gumbel distribution. Let $\h: \R^m \rightarrow \R$ be a multivariate function that satisfies the conditions in Theorem~\ref{theorem:poincare} for each dimension and that $\|\nabla \h(t)\| \le a$. Then whenever $\lambda a \le 1$ the moment generating function is bounded as 
	\begin{align}
	\Lambda(\lambda) \le \frac{1 + \lambda a }{1 - \lambda a }  \cdot \exp \left(\lambda \expe{ \h} \right).
	\end{align}
\end{corollary}
\begin{IEEEproof}
The proof is due to Bobkov and Ledoux~\cite{BobkovL:97exp}. Applying the Poincar\'{e} inequality with $g(t) = \exp(\lambda \h(t)/2)$ implies 
	\begin{align}
	\Lambda(\lambda) - \Lambda(\lambda/2)^2 \le 4 \int \frac{\lambda^2}{4} \exp(\lambda f(t)) \cdot \|\nabla \h(t)\|^2 d \mu(t) \le  a^2 \lambda^2 \Lambda(\lambda).
	\end{align}
Whenever $\lambda^2 a^2 \le 1$ one can rearrange the terms to get the bound $\Lambda(\lambda) \le \Big(1- \lambda^2 a^2  \Big)^{-1} \Lambda(\lambda/2)^2$. Applying this self-reducible bound recursively $k$ times implies
	\begin{align}
	\Lambda(\lambda) &\le \Lambda \left(\frac{\lambda}{2^k} \right)^{2^k} \prod_{i=0}^k \left(1- \frac{\lambda^2 a^2 }{4^i} \right)^{-2^i} \\
	&= \left( 1 + \frac{\lambda \expe{\h}}{2^k} + o(2^k) \right)^{2^k} \prod_{i=0}^k \left(1- \frac{\lambda^2 a^2 }{4^i} \right)^{-2^i},
	\end{align}
where the last line follows from a Taylor expansion of \eqref{eq:mgf}.
Taking $k \rightarrow \infty$ and noting that $(1 + c/2^k)^{2^k} \to e^c$ we obtain the bound
$\Lambda(\lambda) \le \prod_{i=0}^\infty \left(1- \frac{\lambda^2 a^2}{4^i} \right)^{-2^i} \exp(\lambda \expe{\h})$. Applying Lemma \ref{lem:taylor_bound} (see the Appendix for a proof) shows that $\prod_{i=0}^\infty \left(1- \frac{\lambda^2 a^2}{4^i} \right)^{-2^i} \le \frac{1 + \lambda a }{1 - \lambda a }$, which completes the proof.
\end{IEEEproof}

Bounds on the moment generating function generally imply (via the Markov inequality) that the deviation of a random variable from its mean decays exponentially in the number of samples. We apply these inequalities in a setting in which the function $f$ is random and we think of it as a random variable. With some abuse of notation then, we will call $f$ a random variable in the following corollary.

\begin{corollary}[Measure concentration for the Gumbel distribution]
\label{cor:p-concentration}
Consider a random function $f$ that satisfies the same assumptions of Corollary~\ref{cor:p-mgf} almost surely.  Let $\h_1, \h_2, \ldots, \h_M$ be $M$ i.i.d. random variables with the same distribution as $\h$.  Then with probability at least $1-\delta$,
	\begin{equation*}
	\frac{1}{M} \sum_{j=1}^{M} \h_j - \E[\h]
		\leq
		2a \left(1+\sqrt{\frac{1}{2M}\log\frac{1}{\delta}} \right)^2.
	\end{equation*}
	\end{corollary}
	
\begin{IEEEproof}
From the independence assumption, using the Markov inequality, we have that
\begin{align*}
&\prob\left( \sum_{j=1}^M \h_j \leq M \E[\h] + M r\right) \leq \exp(-\lambda M \E[\h] - \lambda M r) \prod_{j=1}^M \E[\exp(\lambda \h_j)].
\end{align*}
Applying Corollary~\ref{cor:p-mgf}, we have, for any $\lambda \leq 1/a$,
\begin{align*}
&\prob\left( \frac{1}{M} \sum_{j=1}^M \h_j \leq \E[\h] + r\right) \le \exp\Big( M( \log(1 + \lambda a) - \log(1 - \lambda a ) - \lambda r) \Big).
\end{align*}
Optimizing over positive $\lambda$ subject to $\lambda \leq 1/a$ we obtain $\lambda = \frac{\sqrt{r - 2a}}{a\sqrt{r}}$ for $r\geq 2a$. Hence, for $r\geq 2a$, the right side becomes
	\begin{align}
	&\exp\left( M\left(2 \tanh^{-1} \left(\sqrt{1 - \frac{2a}{r}}\right) 
		-\frac{r\sqrt{1-\frac{2a}{r}}}{a} \right) \right) \\
	&\quad \leq \exp\left( 2 M\sqrt{\frac{r}{2a} - 1}\left(1-\sqrt{\frac{r}{2a}}\right)\right) \\
	&\quad \leq \exp\left( -2 M\left(\sqrt{\frac{r}{2a}}-1\right)^2\right),
	\end{align}
where the first inequality can be easily verified comparing the derivatives.
Equating the left side of the last inequality to $\delta$ and solving for $r$, we have the stated bound.
\end{IEEEproof}

\subsection{A modified log-Sobolev inequality for log-concave distributions}
\label{sec:sobolev}

In this section we provide complementary measure concentration results that bound the moment generating function $\Lambda(\lambda) = \int \exp(\lambda f(t)) d \mu(t)$ by its expansion (in terms of gradients). Such bounds are known as modified log-Sobolev bounds. We follow the same recipe as previous works~\cite{BobkovL:97exp, Ledoux:01concentration} and use the so-called Herbst argument. Consider the $\lambda$-scaled cumulant generating function of a random function with zero mean, i.e., $\E[\h]=0$:   
	\begin{align}
	K(\lambda) \defeq \frac{1}{\lambda} \log \Lambda(\lambda).
	\label{eq:lcgf}
	\end{align}
First note that by L'H\^{o}pital's rule $K(0) = \frac{ \Lambda'(0) }{ \Lambda(0) } = \E[\h]$, so whenever $\E[\h]=0$ we may represent $K(\lambda)$ by integrating its derivative: $K(\lambda) = \int_0^\lambda K'(\hat \lambda) d \hat \lambda$. Thus to bound the moment generating function it suffices to bound $K'(\lambda) \le \alpha(\lambda)$ for some function $\alpha(\lambda)$. A direct computation of $K'(\lambda)$ translates this bound to    
	\begin{align}
	\lambda \Lambda'(\lambda) - \Lambda(\lambda) \log \Lambda(\lambda) \le \lambda^2 \Lambda(\lambda) \alpha(\lambda).
	\label{eq:logsobdiff}
	\end{align}
The left side of \eqref{eq:logsobdiff} turns out to be the so-called functional entropy $\Ent$, which is not the same as the Shannon entropy~\cite{Ledoux:01concentration}. We calculate the functional entropy of $\lambda f(t)$ with respect to a measure $\mu$:
	\begin{align*}
        \label{eq:s-entropy} 
	\Ent_\mu(\exp(\h)) &\defeq \int \h(t) \cdot \exp(\h(t))  d\mu(t) - \left( \int \exp(\h(t)) d\mu(t) \right) \log \Big(\int \exp(\h(t)) d\mu(t) \Big).
	\end{align*}

In the following we derive a variation of the modified log-Sobolev inequality for log-concave distributions based on a Poincar\'{e} inequality for these distributions. This in turn provides a bound on the moment generating function. This result complements the exponential decay that appears in Section \ref{sec:poincare}. Figure~\ref{fig:mgf} compares these two approaches. 
 
\begin{lemma}
\label{thm:s-mgf}
Let $\mu$ be a measure that satisfies the Poincar\'{e} inequality with a constant $C$, i.e., $\Var_{\mu}(\h) \leq C \int \|\nabla \h(t)\|^2 d\mu(t)$ for any continuous and differentiable almost everywhere function $\h(t)$. Assume that $\int \h(t) d \mu(t) = 0$ and that $\| \nabla \h(t) \| \le a < 2 /\sqrt{C}$. Then 
\[
\Ent_\mu(\exp(\h)) \leq \frac{a^2 C }{2} \left(\frac{2+a\sqrt{C}}{2-a\sqrt{C}}\right)^2 \int \exp(\h(t)) d \mu(t).
\]
\end{lemma} 
\begin{IEEEproof}
First, $z \log z \ge z + 1$. Setting $z = \int \exp(\h) d \mu$ and applying this inequality results in the functional entropy bound $\Ent_\mu(\exp(\h)) \le \int \h(t) \exp(\h(t)) d \mu(t) -  \left(\int \exp(\h(t)) d \mu(t) \right)  + 1$. Rearranging the terms, the right hand side is $\int \left( \h(t) \cdot \exp(\h(t))  - \exp(\h(t))  + 1\right) d \mu(t)$. 
We proceed by using the identity (cf.~\cite{Schaum99}, Equation 17.25.2) of the indefinite integral $\int s \exp(sc) ds = \frac{\exp(sc)}{c}(s-\frac{1}{c})$. Taking into account the limits $[0,1]$ and setting $c=\h(t)$ we get the desired form: $\h(t)^2 \int_0^1 s \exp(s \h(t)) ds = \h(t) \exp(\h(t)) - \exp(\h(t)) + 1$. Particularly,  
	\begin{align}
	\Ent_\mu(\exp(\h)) 
	&\le \int \left(\int_0^1 s \h(t)^2 \exp(s \h(t)) ds \right) d\mu(t) \\
	&=  \lim_{\epsilon \rightarrow 0^+} \int_\epsilon^1 \frac{1}{s} \left( \int  s^2 \h(t)^2 \exp(s \h(t)) d\mu(t) \right) ds.
	\end{align}
The last equality holds by Fubini's theorem. 

Next we use Proposition 3.3 from~\cite{BobkovL:97exp} that applies the Poincar\'{e} inequality to $g(t) \exp(g(t)/2)$ to show that for any function $g(t)$ with mean zero and $\|\nabla g(t)\| \le a < 2/\sqrt{C}$, we have the inequality
	\[
	\int g^2(t) \exp(g(t)) d \mu(t) \le \hat C \int \| \nabla g(t) \|^2 \exp(g(t)) d \mu(t),
	\]
where $\hat C = C ((2+a\sqrt{C}) / (2-a\sqrt{C}))^2$.  Setting $g(t) = s f(t)$ in this inequality satisfies $\|\nabla g(t)\| = s\|\nabla \h(t)\| \le a < 2/\sqrt{C}$. This implies the inequality
\[
    \int  s^2 \h(t)^2 \exp(s \h(t)) d\mu(t) \le s^2 C \left(\frac{2+a\sqrt{C}}{2-a\sqrt{C}}\right)^2 \int \|\nabla \h(t)\|^2 \exp(s \h(t)) d \mu(t).
\]
Using $\|\nabla \h\| \le a$ we obtain the bound 
	\[
	\Ent_\mu(\exp(\h)) \le a^2 C \left(\frac{2+a\sqrt{C}}{2-a\sqrt{C}}\right)^2 \int_0^1 s \left(\int \exp(s \h(t)) d \mu(t) \right) ds.
	\]  

The function $\phi(s)= \int \exp(s \h(t)) d \mu(t)$ is convex in the interval $s \in [0,1]$, so its maximum value is attained at $s=0$ or $s=1$. Also, $\phi(0) = 1$ and $\phi(1)=\int \exp(\h(t)) d \mu(t)$. From Jensen's inequality, and the fact that $\int \h(t) d \mu(t) =0$, we have $\int \exp(\h(t)) d \mu(t) \geq \exp(\int \h(t) d \mu(t)) = 1$. Hence $\phi(1) \geq \phi(0)$. So, we have 
	\begin{align*}
	\int_0^1 s \left(\int \exp(s \h(t)) d \mu(t) \right) ds 
	&\leq \int \exp(\h(t)) d \mu(t) \cdot  \int_0^1 s ds \\
	&= \frac{1}{2} \int \exp(\h(t)) d \mu(t)
	\end{align*}
and the result follows. 
\end{IEEEproof}

The preceding lemma expresses an upper bound on the functional entropy in terms of the moment generating function. Applying this Lemma with the function $\lambda \h$, and assuming that $\|\nabla f\| \le a$, we rephrase this upper bound as $\Ent_\mu(\exp(\lambda f)) \le \Lambda(\lambda) \cdot \frac{\lambda^2 a^2 C}{2} \left(\frac{2+\lambda a\sqrt{C}}{2- \lambda a\sqrt{C}}\right)^2$, where $\Lambda(\lambda)$ is the moment generating function of $f$.  Fitting it to  \eqref{eq:logsobdiff} and \eqref{eq:lcgf} we deduce that 
	\begin{align}
	K'(\lambda) \le \alpha(\lambda) = \frac{a^2 C }{2} \left(\frac{2+\lambda a\sqrt{C}}{2- \lambda a\sqrt{C}}\right)^2.
	\end{align}  
Since the Poincar\'{e} constant of the Gumbel distribution is at most $4$ we obtain its corresponding bound $K'(\lambda) \le  2a^2 \left(\frac{1+\lambda a}{1- \lambda a}\right)^2$. Applying the Herbst argument (cf.~\cite{Ledoux:01concentration}), this translates to a bound on the moment generating function. This result is formalized as follows.

\begin{corollary}
\label{cor:s-mgf}
Let $\mu$ denote the Gumbel measure on $\R$ and let $f: \R^m \rightarrow \R$ be a multivariate function that satisfies the conditions in Lemma~\ref{lemma:poincare} for each dimension. Also, assume that $\norm{ \nabla f(t) } \le a$. 
Then whenever $\lambda a \le 1$  the moment generating function is bounded as
	\begin{align*}
	\Lambda(\lambda) \leq \beta(\lambda) \, \exp{ \left(\lambda \expe{ \h} \right)},
	\end{align*}
where $\beta(\lambda) = \exp\left(2a^2\lambda^2  \frac{5-\lambda a}{1-\lambda a} + 8 a \lambda \log (1-\lambda a)\right)$.
\end{corollary}
\begin{IEEEproof}
We apply Lemma \ref{thm:s-mgf} to the function $\hat \h(t) = \h(t) - \E[\h]$ which has zero mean. Thus 
	\[
	K'(\lambda)=\frac{\Ent_\mu (\exp(\lambda \hat \h))}{\lambda^2 \Lambda(\lambda)} \leq 2a^2 \left(\frac{1+\lambda a}{1- \lambda a}\right)^2.
	\]
Recalling that $K(0)=0$ we derive $K(\lambda) = \int_0^\lambda K'(\hat \lambda) d \hat \lambda$. Using the bound on $K'(\lambda)$ we obtain 
	\[
	K(\lambda) \le 2a^2 \int_0^\lambda \left(\frac{1+\hat \lambda a}{1- \hat \lambda a}\right)^2 d \hat \lambda.
	\]
A straight forward verification of the integral implies that 
	\begin{align*}
	K(\lambda) &\le 2a^2 \left[ \frac{4\log(1-a\lambda)}{a} + \frac{4}{a(1-a\lambda)}+\lambda - \frac{4}{a}\right] \\
	&= 2a^2 \left[ \frac{4\log(1-a\lambda)}{a} + \frac{4 \lambda}{1-a\lambda}+\lambda \right].
	\end{align*}
Now, from the definition of $K(\lambda)$ and the one of $\beta(\lambda)$, this implies $\log \E[\exp(\lambda \hat \h)] \leq \log \beta(\lambda)$. 
\end{IEEEproof}

\subsection{Evaluating measure concentration}

\begin{figure}
\centering
\subfloat[]{
	\includegraphics[width=0.315\linewidth]{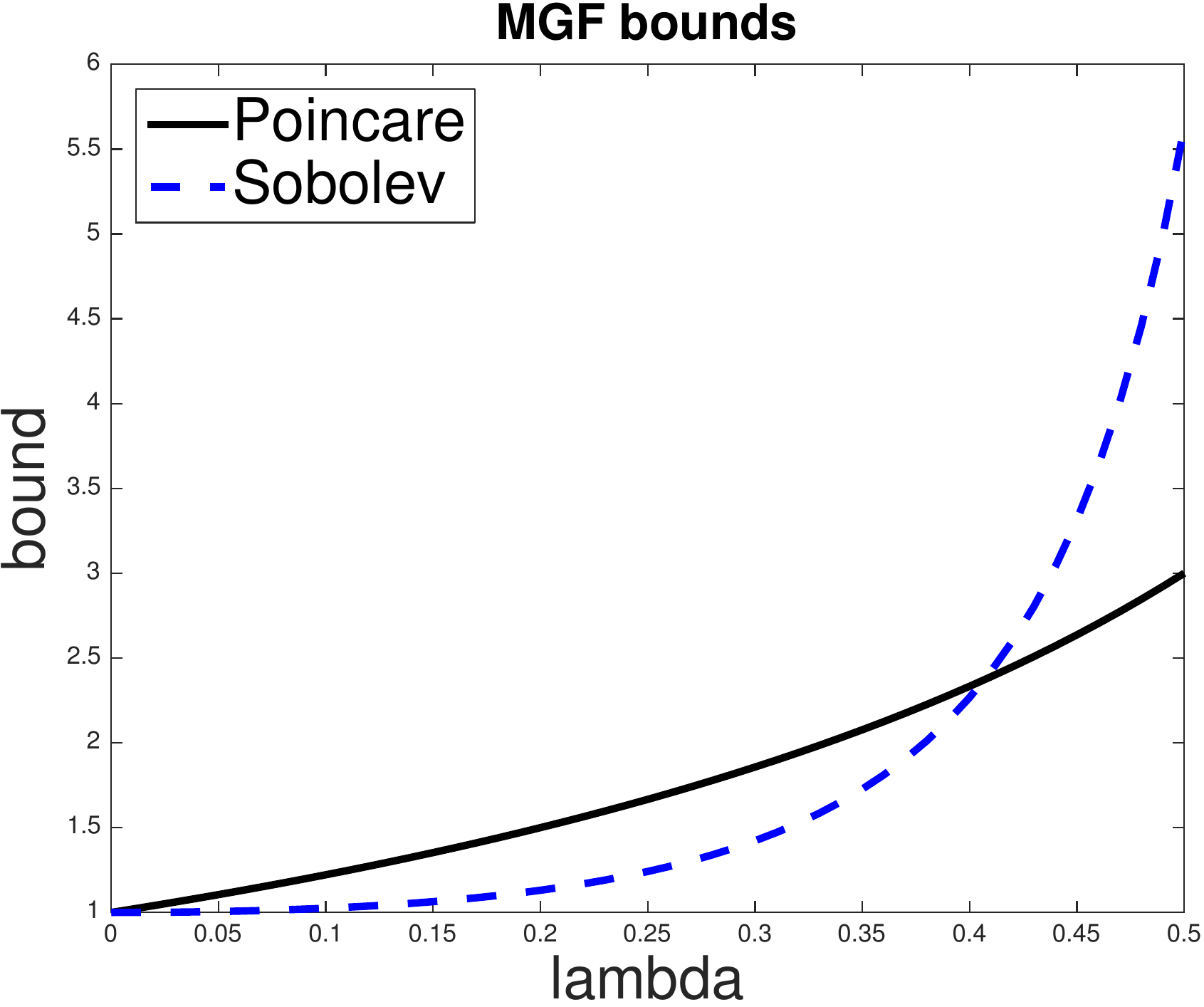}
	\label{fig:mgfbounds}
	}%
\subfloat[]{
	\includegraphics[width=0.315\linewidth]{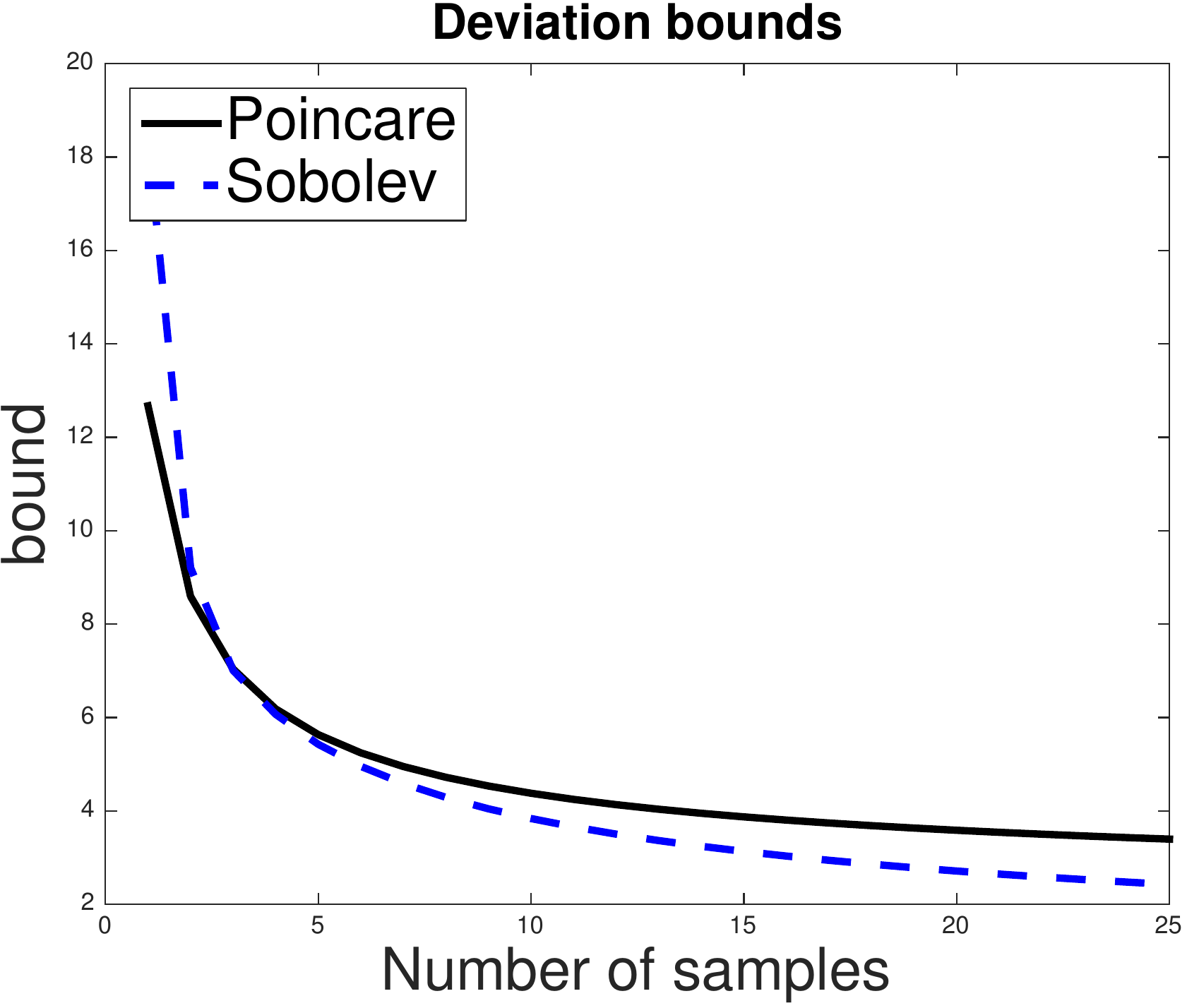}
	\label{fig:devboundsM}
}%
\subfloat[]{
	\includegraphics[width=0.315\linewidth]{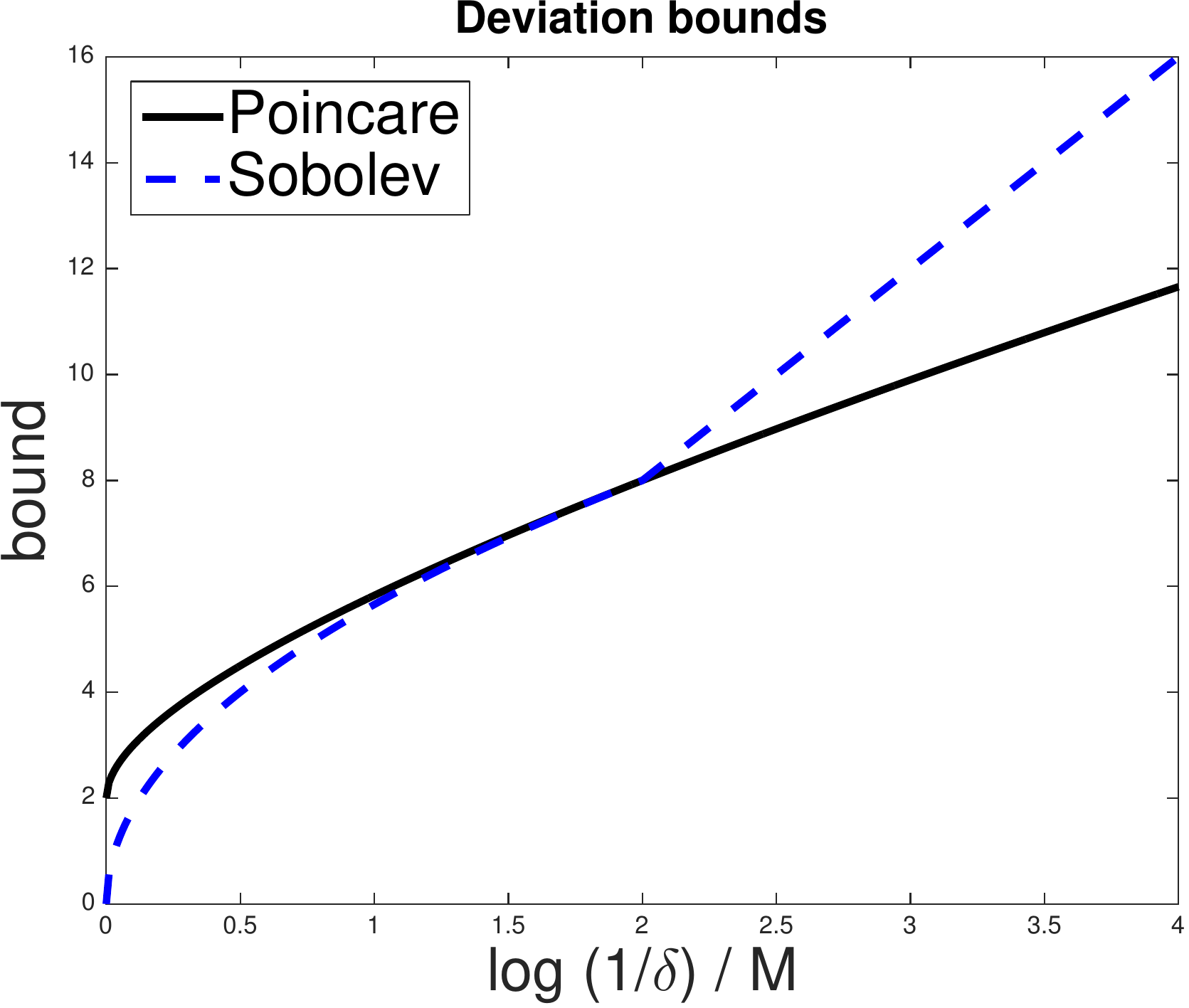}
	\label{fig:devboundsdM}
}
\caption{
Comparing the measure concentration bounds that are attained by the Poincar\'{e} and modified log-Sobolev inequalities for $a=1$.
Figure (\ref{fig:mgfbounds}): the moment generating functions bounds that are attained by the Poincar\'{e} inequality in Corollary \ref{cor:p-mgf} and the modified log-Sobolev inequality in Corollary \ref{cor:s-mgf}. The plots show that respective functions $\alpha(\lambda) =  \frac{1 + \lambda a }{1 - \lambda a }$ and $\beta(\lambda)$ that appear in these bounds.
Figure (\ref{fig:devboundsM}): the deviation bounds, of the sampled average from its mean, that are attained by the Poincar\'{e} inequality in Corollary \ref{cor:p-concentration} and the modified log-Sobolev inequality in Corollary \ref{cor:cor2} when $\delta = 0.1$ and the number of samples $M=1,2,\ldots,100$.
Figure (\ref{fig:devboundsdM}): the deviation bounds, of the sampled average from its mean, that are attained by the Poincar\'{e} inequality in Corollary \ref{cor:p-concentration} and the modified log-Sobolev inequality in Corollary \ref{cor:cor2}, as a function for  $\log(1/\delta)/M$ that ranges between $[0,2]$. 
\label{fig:mgf}}
\end{figure}

The above bound is tighter than our previous bound in Theorem 3 of~\cite{Orabona14}. In particular, the bound in Corollary \ref{cor:s-mgf} does not involve $\|f(t)\|_\infty$. It is interesting to compare $\beta(\lambda)$ in the above bound to the one that is attained directly from Poincar\'{e} inequality in Corollary \ref{cor:p-mgf}, namely $\Lambda(\lambda) \le \alpha(\lambda)  \cdot \exp{ (\lambda \expe{ \h})}$ where $\alpha(\lambda) =  \frac{1 + \lambda a }{1 - \lambda a }$. Both $\alpha(\lambda), \beta(\lambda)$ are finite in the interval $0 \le \lambda < 1/a$ although they behave differently at the their limits. Particularly, $\alpha(0)=1$ and $\beta(0) = 1$. On the other hand, $\alpha(\lambda) < \beta(\lambda)$ for $\lambda \rightarrow 1$. This is illustrated in Figure \ref{fig:mgf}. 

With this Lemma we can now upper bound the error in estimating the average $\E[\h]$ of a function $\h$ of $m$ i.i.d. Gumbel random variables by generating $M$ independent samples of $\h$ and taking the sample mean. We again abuse notation to think of $\h$ as a random variable itself.

\begin{corollary}[Measure concentration via log-Sobolev inequalities]
\label{cor:cor2}
Consider a random function $f$ that satisfies the same assumptions as Corollary~\ref{cor:s-mgf} almost surely. Let $\h_1, \h_2, \ldots, \h_M$ be $M$ i.i.d. random variables with the same distribution as $\h$.  Then with probability at least $1-\delta$,
	\begin{equation*}
	\frac{1}{M} \sum_{j=1}^{M} \h_j - \E[\h]
		\le a 
		\max\left(\frac{4}{M} \log\frac{1}{\delta},\sqrt{\frac{32}{M} \log\frac{1}{\delta}}\right).
	\end{equation*}
	\end{corollary}

\begin{IEEEproof}
From the independence assumption, using the Markov inequality, we have that
\begin{align*}
\prob\left( \sum_{j=1}^M \h_j \leq M \E[\h] + M r\right) \leq \exp(-M \lambda \E[\h]-M r \lambda) \prod_{j=1}^M \E[\exp(\lambda \h_j)].
\end{align*}
We use the elementary inequality $\log(1-x) \leq \frac{-2x}{2-x}$ and Corollary~\ref{cor:s-mgf}, to have
\begin{align*}
&\prob\left( \frac{1}{M} \sum_{j=1}^M \h_j \leq \E[\h] + r\right) \le \exp\left(M \left( 2 a^2 \lambda^2 \frac{a^2 \lambda^2+a \lambda+2}{(1-a \lambda)(2-a \lambda)}-\lambda r \right)\right).
\end{align*}
For\footnote{The constants are found in order to have the junction of curve to approximately lie on the Poincare curve in Figure~\ref{fig:devboundsdM}.} any $|\lambda| \leq \frac{13}{25 a}$, we have that $2\frac{a^2 \lambda^2+a \lambda+2}{(1-a \lambda)(2-a \lambda)} \leq 8$.
Hence, for any $|\lambda| \leq \frac{13}{25 a}$, we have that
\begin{align*}
&\prob\left( \frac{1}{M} \sum_{j=1}^M \h_j \leq \E[\h] + r\right) \le \exp\left(M \left( 8 a^2 \lambda^2 -\lambda r\right)\right).
\end{align*}

Optimizing over $\lambda$ subject to $|\lambda| \leq \frac{13}{25 a}$ we obtain
\begin{align*}
\exp\left(M \left(8 a^2 \lambda^2 -\lambda r\right)\right)\leq \exp\left(-M\min\left(\frac{r}{4 a}, \frac{r^2}{32 a^2}\right)\right).
\end{align*}
Equating the left side of the last inequality to $\delta$ and solving for $r$, we have the stated bound.
\end{IEEEproof}

\subsection{Application to MAP perturbations}
\label{sec:bounds}
 
The derived bounds on the moment generating function imply the concentration of measure of our high-dimensional inference algorithms that we use both for sampling (in Section \ref{sec:unbiased}) and to estimate prediction uncertainties or entropies (in Section \ref{sec:entropy}). The relevant quantities for our inference algorithms are the expectation of randomized max-solvers $F(\gamma) = \max_{\x} \{\theta(\x) + \sum_\alpha \gamma_\alpha(\x_\alpha) \}$ as well as the expectation of the maximizing perturbations  $F(\gamma) = \sum_\alpha \gamma_\alpha(\x_\alpha^{\gamma})$ for which $x^{\gamma}  = \argmax_{\x}  \{\theta(\x) + \sum_\alpha \gamma_\alpha(\x_\alpha) \}$, as in Theorem \ref{theorem:bound}. 

To apply our measure concentration results to perturb-max inference we calculate the parameters in the bound given by the Corollary~\ref{cor:p-mgf} and Corollary~\ref{cor:s-mgf}. The random functions $F(\gamma)$ above  are functions of $m \defeq \sum_{\alpha \in {\cal A}} |\cX_\alpha|$ i.i.d. Gumbel random variables. The (sub)gradient of these functions is structured and points toward the $\gamma_\alpha(\x_\alpha)$ that corresponding to the maximizing assignment in $\Xrmap$, that is
	\begin{align*}
	\frac{\partial \mvh(\gamma)}{\partial \gamma_\alpha(x_\alpha)}
		=
		\begin{cases}
		1 & \textrm{if\ } x_\alpha \in \Xrmap \\
		0 & \textrm{otherwise}
		\end{cases}
	\end{align*}
Thus the gradient satisfies $\norm{ \nabla \mvh }^2 = |{\cal A}|$ almost everywhere, so $a^2=|{\cal A}|$. Suppose we sample $M$ i.i.d. random variables $\gamma_1, \gamma_2, \ldots, \gamma_M$ with the same distribution as $\gamma$ and denote their respective random values by $F_j \stackrel{def}{=} F(\gamma_j)$.  We estimate their deviation from the expectation by $\frac{1}{M} \sum_{i=1}^{M} \mvh_j - E[\mvh]$. Applying Corollary \ref{cor:p-mgf} to both $\mvh$ and $-\mvh$ we get the following double-sided bound with probability $1 - \delta$:
	\begin{equation*}
	\frac{1}{M} \sum_{j=1}^{M} \mvh_j - \E[\mvh]
		\leq
		2\sqrt{|{\cal A}|}  \left( 1+\sqrt{\frac{1}{2M} \log\frac{2}{\delta}}\right)^2.
	\end{equation*}
Applying Corollary \ref{cor:s-mgf} to both $\mvh$ and $-\mvh$ we get the following double-sided bound with probability $1 - \delta$:
	\begin{equation*}
	\frac{1}{M} \sum_{j=1}^{M} \mvh_j - \E[\mvh]
		\leq
                 \sqrt{|{\cal A}|} \max\left(\frac{4}{M} \log\frac{2}{\delta},\sqrt{\frac{32}{M} \log\frac{2}{\delta}}\right).
	\end{equation*}
Clearly, the concentration of perturb-max inference is determined by the best concentration of these two bounds.

\section{Empirical Evaluation}
\label{sec:experiments}

\begin{figure*}
\centering
\subfloat[][]{
	\includegraphics[trim=30 30 30 25pt,clip,width=0.24\textwidth]{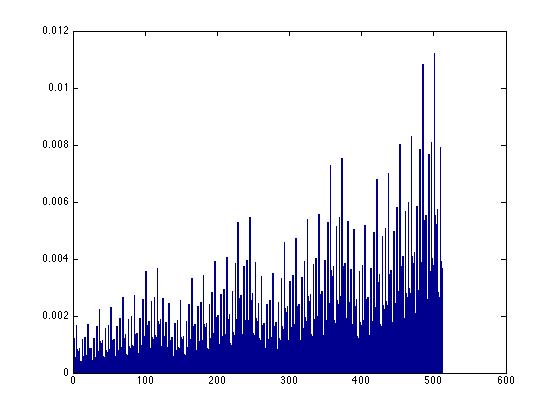}
}%
\subfloat[][]{
	\includegraphics[trim=30 30 30 25pt,clip,width=0.24\textwidth]{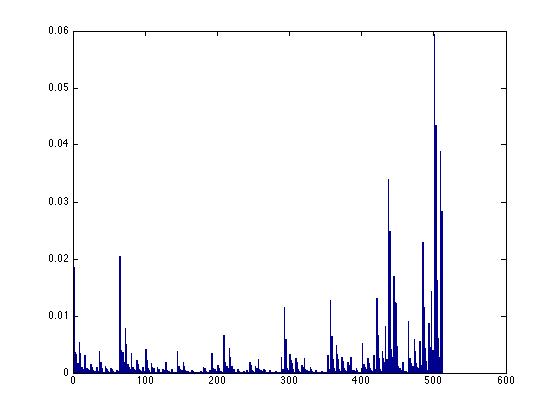}
}%
\subfloat[][]{
	\includegraphics[trim=30 30 30 25pt,clip,width=0.24\textwidth]{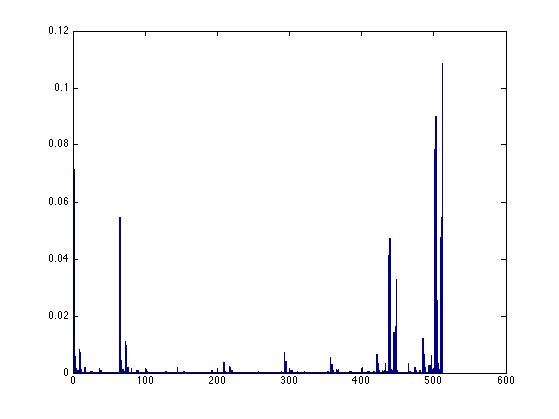}
}%
\subfloat[][]{
	\includegraphics[trim=30 30 30 25pt,clip,width=0.24\textwidth]{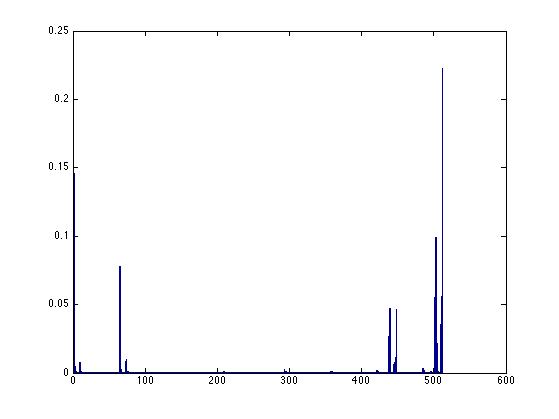}
}%
\\
\subfloat[$\theta_{i,j}=0$]{
	\includegraphics[trim=30 30 30 25pt,clip,width=0.24\textwidth]{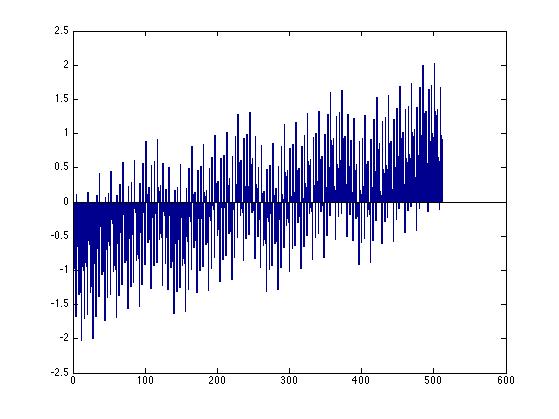}
}%
\subfloat[$\theta_{i,j} \in [0,1)$]{
	\includegraphics[trim=30 30 30 25pt,clip,width=0.24\textwidth]{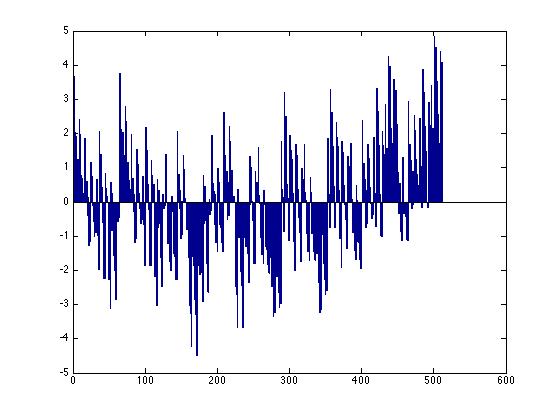}
}%
\subfloat[$\theta_{i,j} \in [0,2)$]{
	\includegraphics[trim=30 30 30 25pt,clip,width=0.24\textwidth]{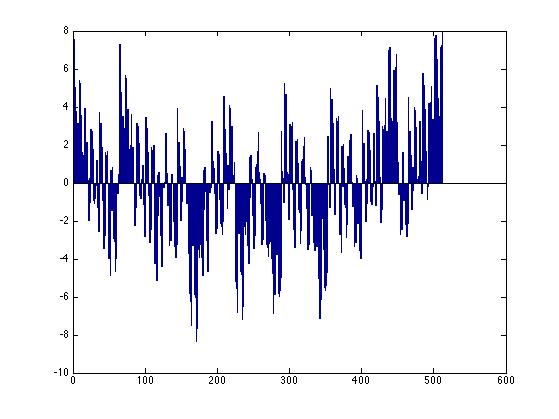}
}%
\subfloat[$\theta_{i,j} \in [0,3)$]{
	\includegraphics[trim=30 30 30 25pt,clip,width=0.24\textwidth]{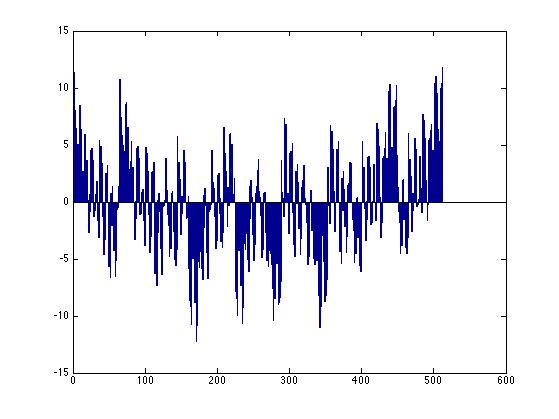}
}%
\caption{
The probability (top row) and energy (bottom row) landscapes for all $512$ configurations in a $3 \times 3$ spin glass system with strong local field,  $\theta_i \in [-1,1]$. When $\theta_{i,j}=0$ the system is independent and one can observe the block pattern. As the coupling potentials get stronger the landscape get more ragged. By zooming one can see the ragged landscapes  throughout the space, even for negligible configurations, which affect many local approaches. The random MAP perturbation directly targets the maximal configurations, thus performs well in these settings. 
}
\label{fig:landscape}
\end{figure*}

Statistical inference of high dimensional structures is closely related to estimating the partition function. Our proposed inference algorithms, both for sampling and inferring the entropy of high-dimensional structures, are derived from an alternative interpretation of the partition function as the expected value of the perturb-max value. We begin our empirical validation by computing the upper and lower bounds for the partition function computed as the expected value of a max-function. We then show empirically that the perturb-max algorithm for sampling from the Gibbs distribution has a sub-exponential computational complexity. Subsequently, we evaluate the properties of the perturb-max entropy bounds. Also, we explore the deviation of the sample mean of the perturb-max value from its expectation by deriving new measure concentration inequalities. Lastly, we use MAP perturbations inference as a sub-procedure for supervised learning binary image denoising (spin glass model) and demonstrate its power over structured-SVMs.

We evaluate our approach on spin glass models, where each variable $x_i$ represents a spin, namely $x_i \in \{-1,1\}$. Each spin has a local field parameter $\theta_i$ which correspond to the local potential function $\theta_i(x_i) = \theta_i x_i$. The parameter $\theta_i$ represents data signal, which in the spin model is the preference of a spin to be positive or negative. Adjacent spins interact with couplings $\theta_{i,j} (x_i x_j) = \theta_{i,j} x_i x_j$. Whenever the coupling parameters are positive the model is called attractive because adjacent variables give higher values to positively correlated configurations. The potential function of a spin glass model is then 
	\[
	\theta(x_1,x_2,\ldots,x_n) = \sum_{i \in V} \theta_i x_i  + \sum_{(i,j) \in E} \theta_{i,j} x_i x_j.
	\]
In our experiments we consider adjacencies of a grid-shaped model. We used low dimensional random perturbations $\gamma_i(x_i)$ since such perturbations do not affect the complexity of the MAP solver. 

Evaluating the partition function is challenging when considering strong local field potentials and coupling strengths. The corresponding energy landscape is ragged, and characterized by a relatively small set of dominating configurations. An example of these energy and probability landscapes are presented in Figure \ref{fig:landscape}. 

\begin{figure}
\centering
\includegraphics[width=8.5cm]{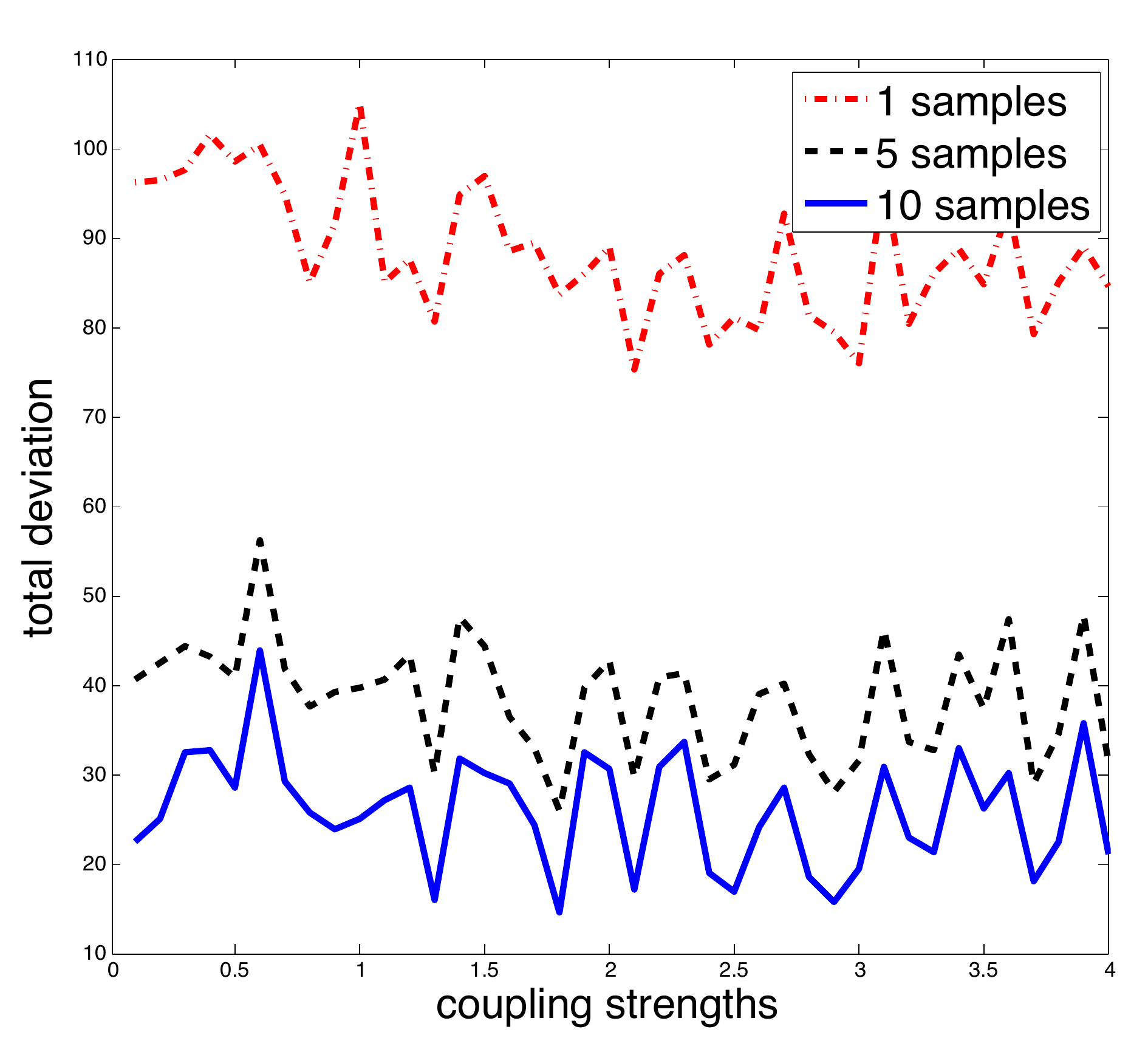}
\caption{ Error of the sample mean versus coupling strength for $100 \times 100$ spin glass models. The local field parameter $\theta_i$ is chosen uniformly at random from $[-1, 1]$ to reflect high signal. With only $10$ samples one can estimate the expectation well. \label{fig:samplemean}}
\end{figure}

\begin{figure*}
	\centering
	\subfloat{
		\centering
		\includegraphics[width=0.32\textwidth]{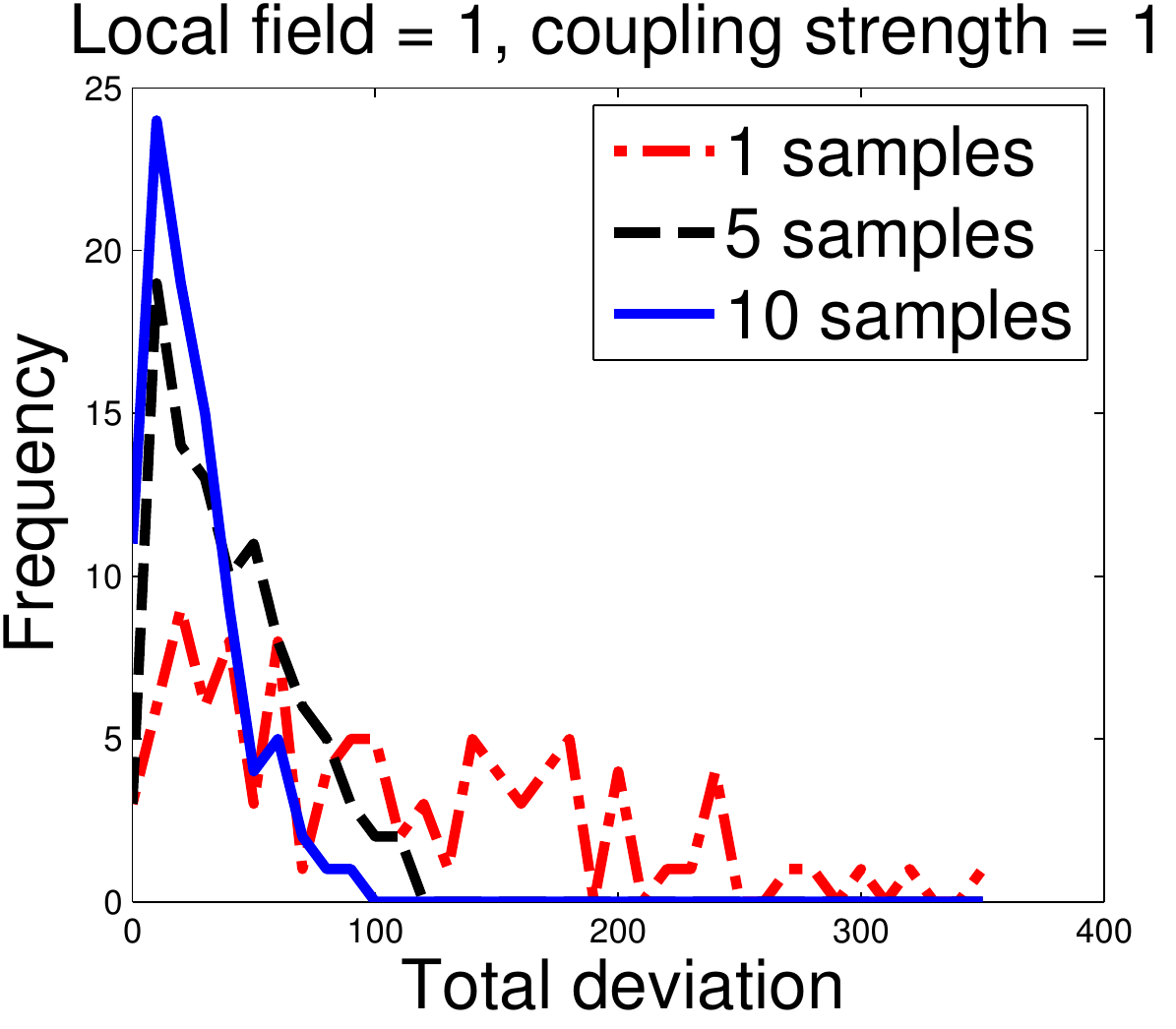}
	}%
	\subfloat{
		\centering
		\includegraphics[width=0.32\textwidth]{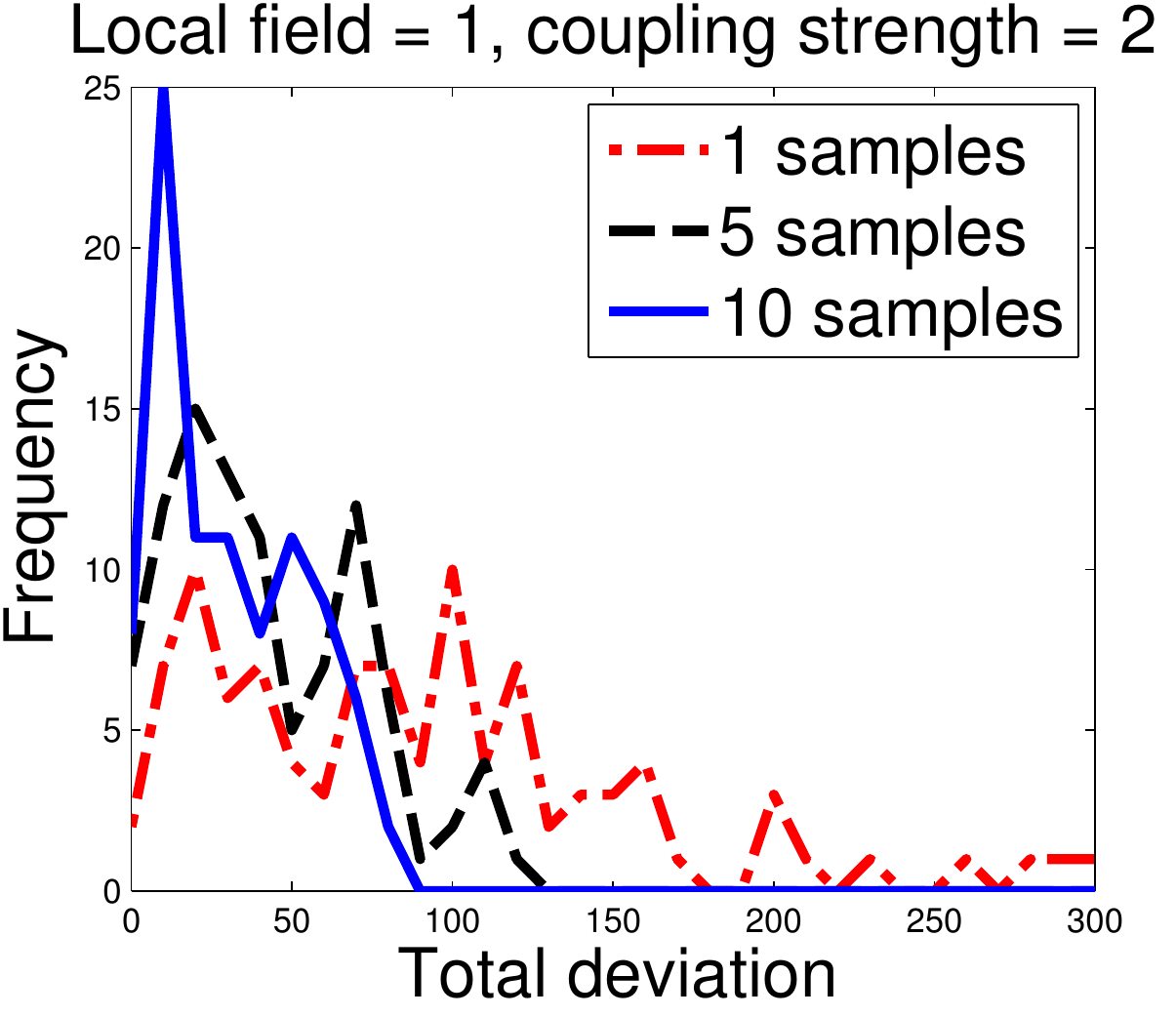}
	}%
	\subfloat{
		\centering
		\includegraphics[width=0.32\textwidth]{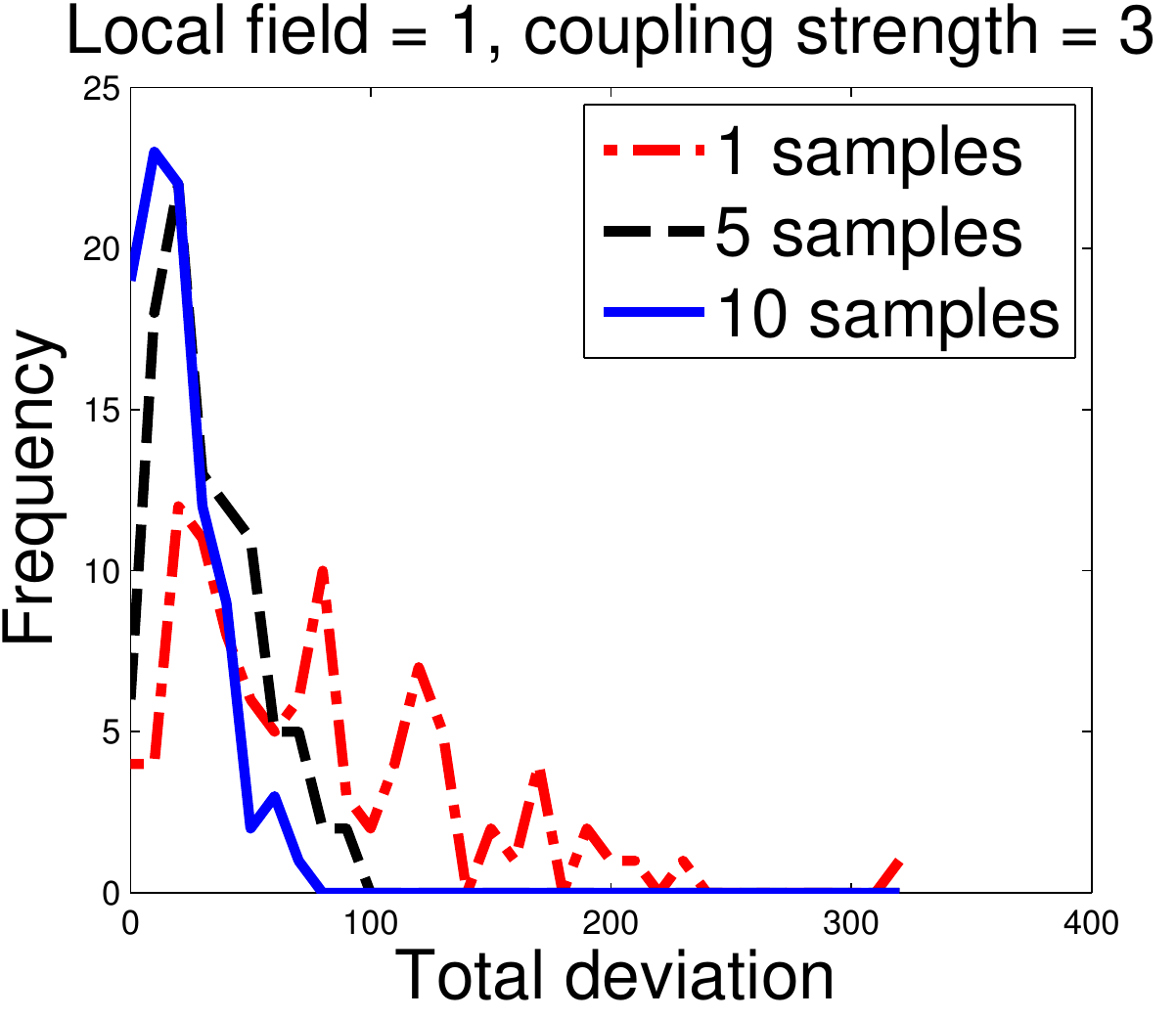}
	}%
\caption{ 
Histogram that shows the decay of random MAP values, i.e., the number of times that the sample mean has error more than $r$ from the true mean. These histograms are evaluated on $100 \times 100$ spin glass model with high signal $\theta_i \in [-1,1]$ and various coupling strengths. One can see that the decay is indeed exponential for every $M$, and that for larger $M$ the decay is faster.
\label{fig:spin-glass}}
\end{figure*}

First, we compared our bounds to the partition function on $10 \times 10$ spin glass models. For such comparison we computed the partition function exactly using dynamic programming (the junction tree algorithm). The local field parameters $\theta_i$ were drawn uniformly at random from $[-f,f]$, where $f \in \{0.1, 1\}$ reflects weak and strong data signal. The parameters $\theta_{i,j}$ were drawn uniformly from $[0,c]$ to obtain attractive coupling potentials. Attractive potentials are computationally favorable as their MAP value can be computed efficiently by the graph-cuts algorithm~\cite{Boykov01}. First, we evaluate an upper bound in (\ref{basic:upper}) that holds in expectation with perturbations $\gamma_i(x_i)$. The expectation was computed using $100$ random MAP perturbations, although very similar results were attained after only $10$ perturbations, e.g., Figure \ref{fig:samplemean} and Figure \ref{fig:spin-glass}. We compared this upper bound with the sum-product form of tree re-weighted belief propagation with uniform distribution over the spanning trees~\cite{Wainwright-05upper}. We also evaluate our lower bound that holds in probability and requires only a single MAP prediction on an expanded model, as described in Corollary \ref{corollary:p-lower}. We estimate our probable bound by expanding the model to $1000 \times 1000$ grids, setting the discrepancy $\epsilon$ in Corollary \ref{corollary:p-lower}
 to zero.\footnote{The empirical results show that even with $\epsilon=0$, it is still a lower bound is tighter, with high probability, which may hint that there is a better analysis for this tight bound.}  We compared this lower bound to the belief propagation algorithm, that provides the tightest lower bound on attractive models~\cite{Sudderth08,Ruozzi12,Weller14}.  We computed the signed error (the difference between the bound and $\log Z$), averaged over $100$ spin glass models, see Figure \ref{fig:spin-glass}. One can see that the probabilistic lower bound is the tightest when considering the medium and high coupling domain, which is traditionally hard for all methods. Because the bound holds only with high probability probability it might generate a (random) estimate which is not a proper lower bound. We can see that on average this does not happen. Similarly, our perturb-max upper bound is better than the tree re-weighted upper bound in the medium and high coupling domain. In the attractive setting, both our bounds use the graph-cuts algorithm and were therefore considerably faster than the belief propagation variants. Finally, the sum-product belief propagation lower bound performs well on average, but from the plots one can observe that its variance is high. This demonstrates the typical behavior of belief propagation: it finds stationary points of the non-convex Bethe free energy and therefore works well on some instances but does not converge or attains bad local minima on others.

\begin{figure*}
	\centering
	\subfloat{
		\centering
		\includegraphics[width=0.48\textwidth]{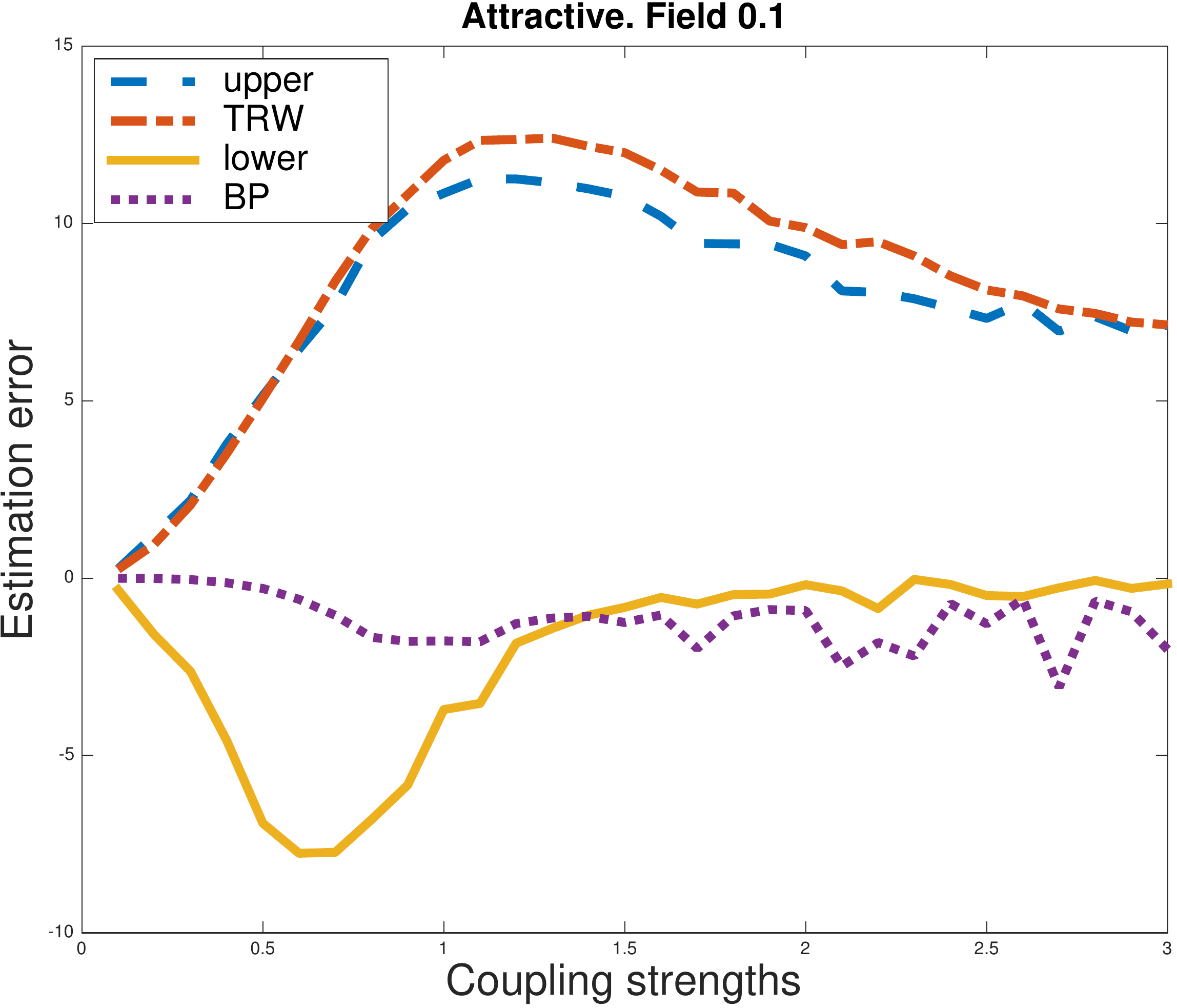} 
	}%
	\subfloat{
		\centering
		\includegraphics[width=0.48\textwidth]{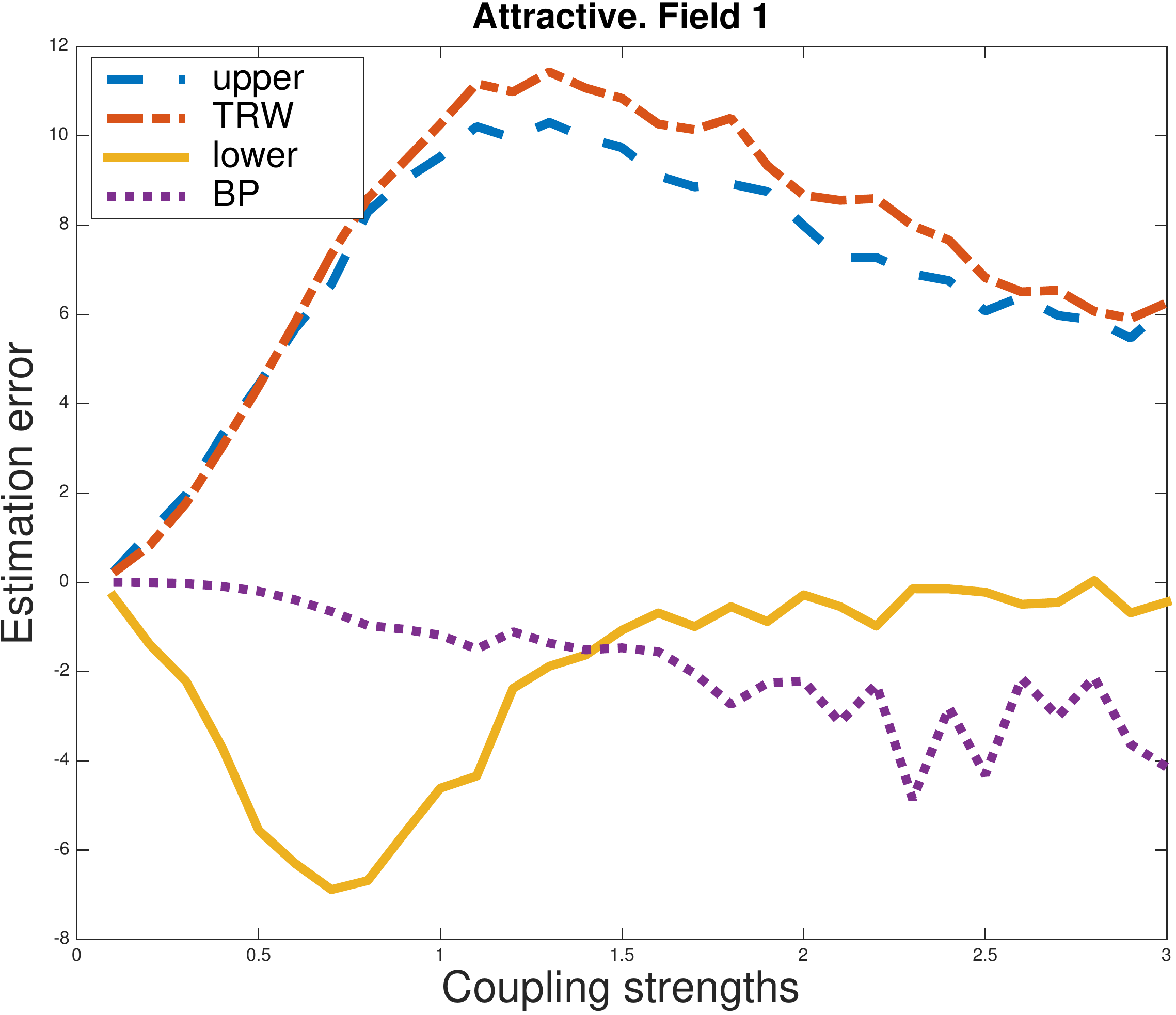}
	}%
\caption{
The attractive case. The (signed) difference of the different bounds and the log-partition function. These experiments illustrate our bounds on $10 \times 10$ spin glass model with weak and strong local field potentials and attractive coupling potentials. The plots below zero are lower bounds and plots above zero are upper bounds. We compare our upper bound \eqref{basic:upper} with the tree re-weighted upper bound. We compare our lower bound (Corollary \ref{corollary:p-lower}) with the belief propagation result, whose stationary points are known to be lower bounds to the log-partition function for attractive spin-glass models.
}    
\label{fig:spin-glass-attractive}
\end{figure*}

We also compared our bound in the mixed case, where the coupling potentials may either be attractive or repulsive, namely $\theta_{ij} \in [-c,c]$. Recovering the MAP solution in the mixed coupling domain is harder than the attractive domain. Therefore we could not test our lower bound in the mixed setting as it relies on expanding the model. We also omit the comparison to the sum-product belief propagation since it is no longer a lower bound in this setting. We evaluate the MAP perturbation value using MPLP~\cite{Sontag-uai08}. One can verify that qualitatively the perturb-max upper bound is significantly better than the tree re-weighted upper bound. Nevertheless it is significantly slower as it relies on finding the MAP solution, a harder task in the presence of mixed coupling strengths.  
 
\begin{figure*}
	\centering
	\subfloat{
		\centering
		\includegraphics[width=0.48\textwidth]{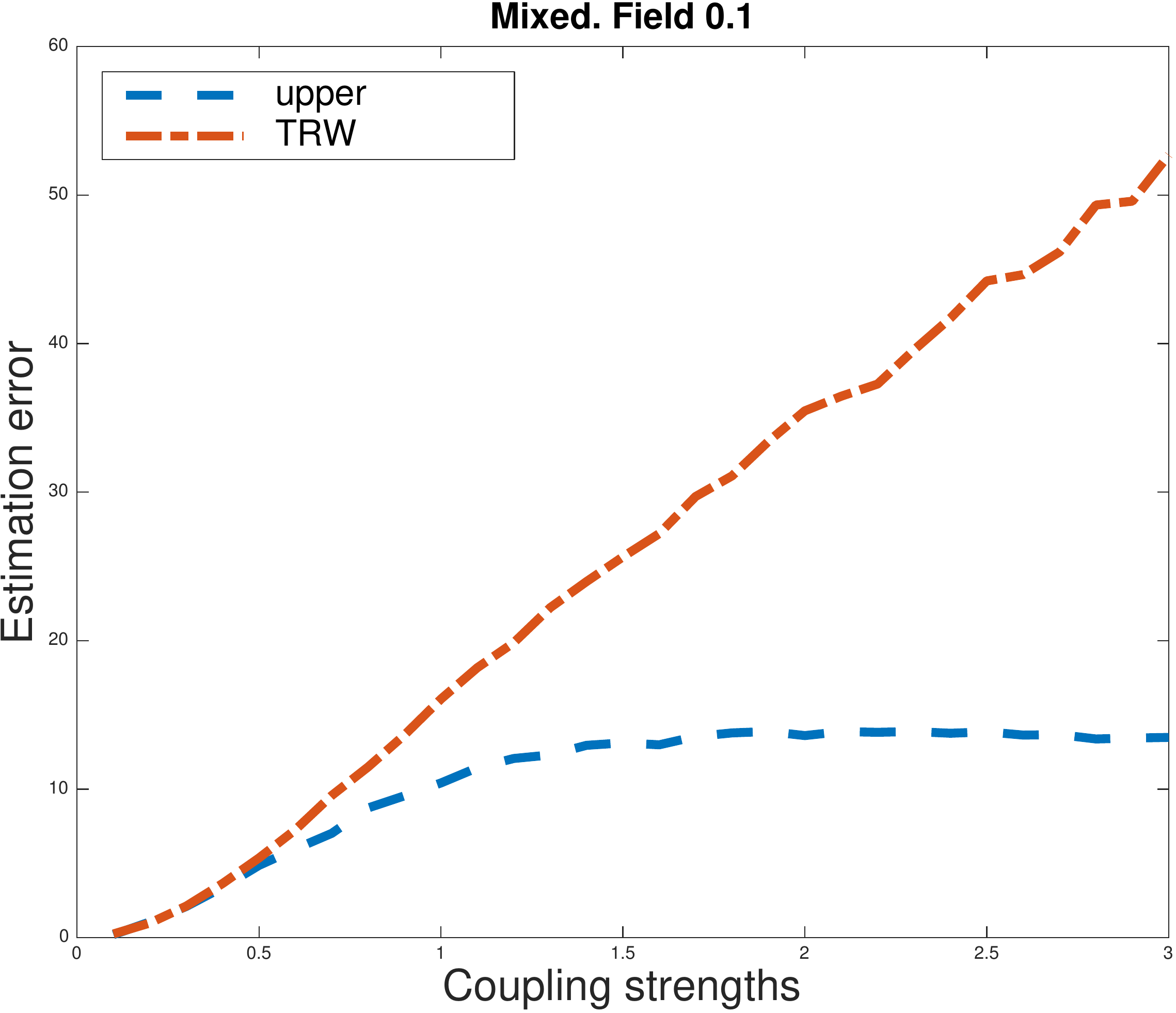} 
	}%
	\centering
	\subfloat{
		\centering
		\includegraphics[width=0.48\textwidth]{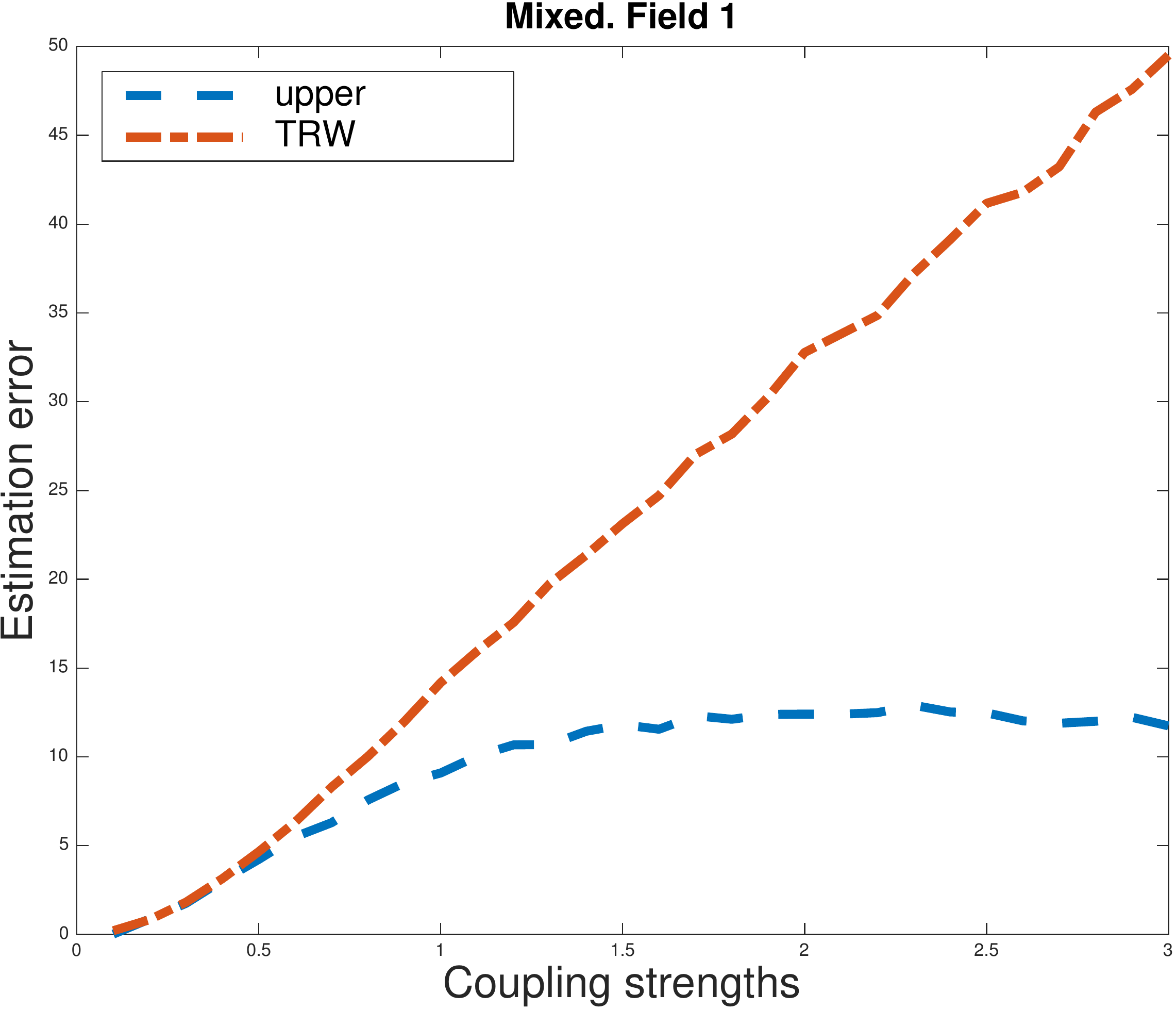}
	}%
\caption{
The (signed) difference of the different bounds and the log-partition function. These experiments illustrate our bounds on $10 \times 10$ spin glass model with weak and strong local field potentials and mixed coupling potentials. We compare our upper bound \eqref{basic:upper} with the tree re-weighted upper bound.
\label{fig:spin-glass-mixed}
}
\end{figure*}

\begin{figure}
\centering
\includegraphics[width=0.49\textwidth]{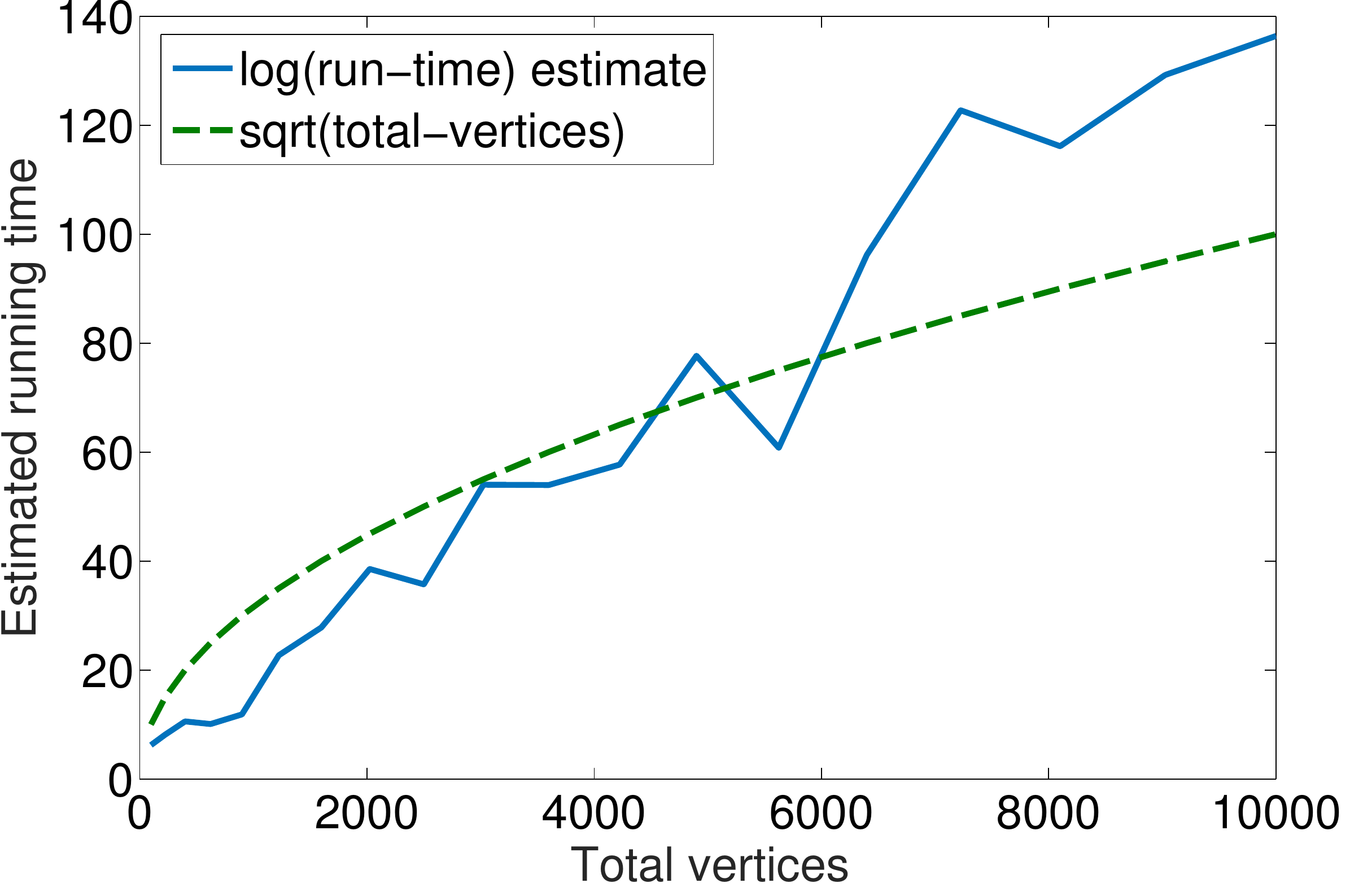} 
\caption{
Estimating our unbiased sampling procedure complexity on spin glass models of varying sizes, ranging from $10 \times 10$ spin glass models to $100 \times 100$ spin glass models. The running time is the difference between our upper bound in (\ref{basic:upper}) and the log-partition function. Since the log-partition function cannot be computed for such a large scale model, we replaced it with its lower bound in Corollary \ref{corollary:p-lower}. 
}    
\label{fig:unbiased}
\end{figure}

Next, we evaluate the computational complexity of our sampling procedure. Section \ref{sec:unbiased} describes an algorithm that generates unbiased samples from the full Gibbs distribution. For spin glass models with strong local field potentials, it is well-known that one cannot produce unbiased samples from the Gibbs distributions in polynomial time~\cite{Jerrum93, Goldberg07, Goldberg12}. Theorem \ref{theorem:unbiased} connects the computational complexity of our unbiased sampling procedure to the gap between the log-partition function and its upper bound in  \eqref{basic:upper}. We use our probable lower bound to estimate this gap on large grids, for which we cannot compute the partition function exactly. Figure \ref{fig:unbiased} suggests that in practice, the running time for this sampling procedure is sub-exponential.

\begin{figure}
	\centering
	\subfloat{
		\centering
		\includegraphics[width=0.452\textwidth]{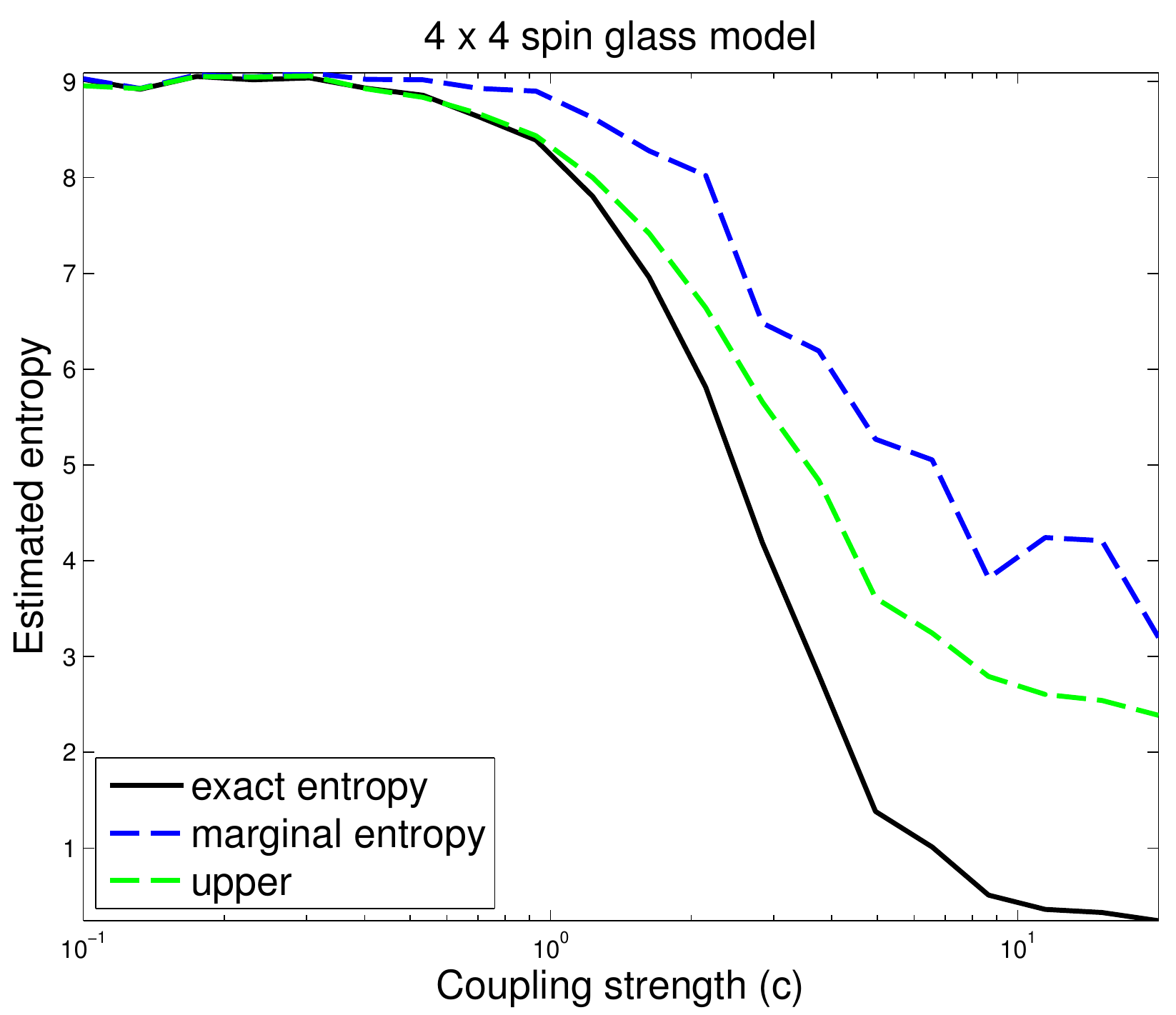} 
	}%
	\centering
	\subfloat{
		\centering
		\includegraphics[width=0.48\textwidth]{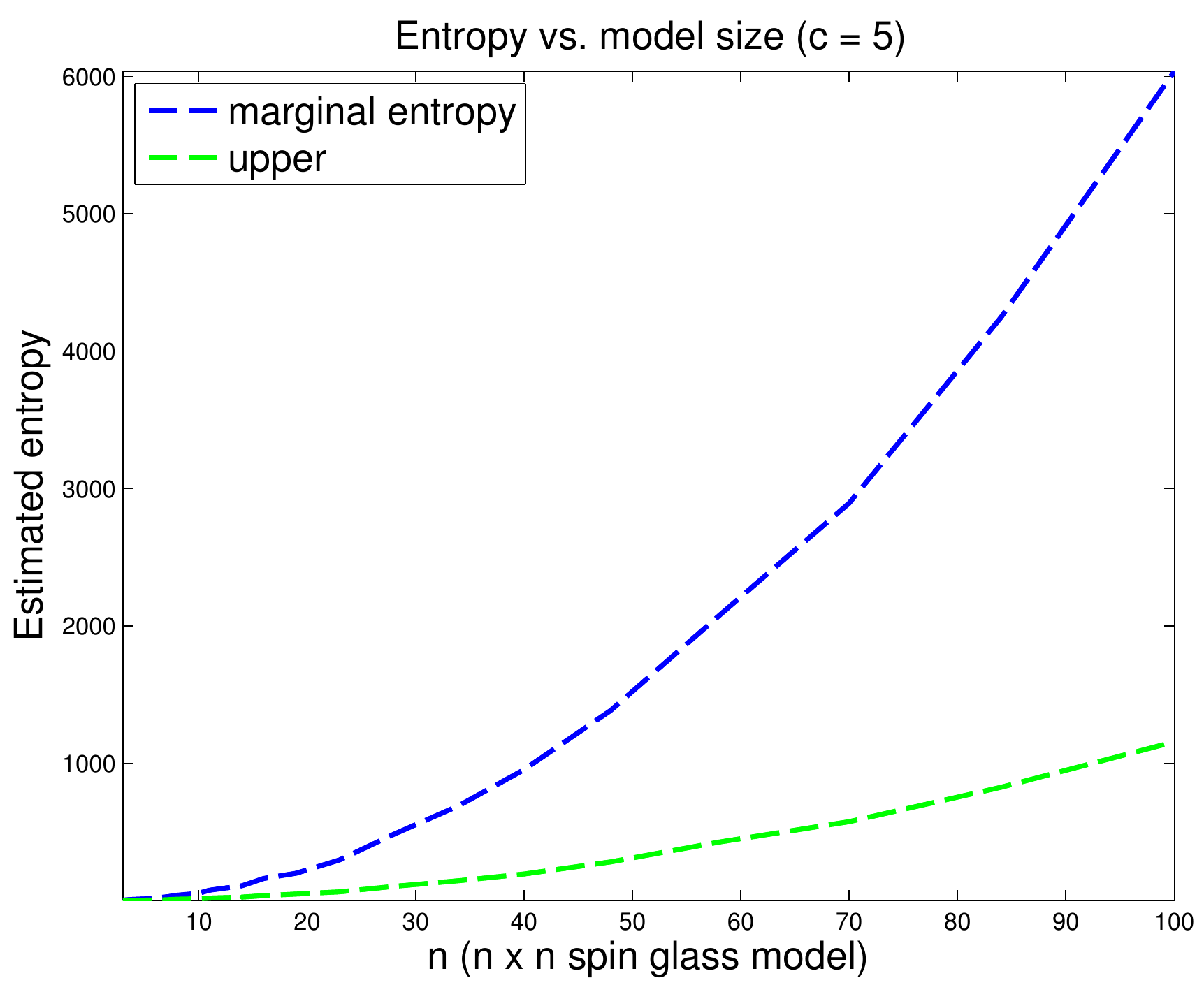} 
	}%
\caption{
Estimating our entropy bounds (in Section \ref{sec:entropy}) while comparing them to the true entropy and the marginal entropy bound. Left: comparison on small-scale spin models. Right: comparison on large-scale spin glass models.
}    
\label{fig:entropy}
\end{figure}

Next we estimate our upper bounds for the entropy of perturb-max probability models that are described in Section \ref{sec:entropy}. We compare them to marginal entropy bounds $H(p) \le \sum_i H(p_i)$, where $p_i(x_i) = \sum_{x \setminus x_i} p(x)$ are the marginal probabilities~\cite{Cover12}. Unlike the log-partition case which relates to the entropy of Gibbs distributions, it is impossible to use dynamic programming to compute the entropy of perturb-max models. Therefore we restrict ourselves to a $4 \times 4$ spin glass model to compare these upper bounds as shown in Figure \ref{fig:entropy}. One can see that the MAP perturbation upper bound is tighter than the marginalization upper bound in the medium and high coupling strengths. We can also compare the marginal entropy bounds and the perturb-max entropy bounds to arbitrary grid sizes without computing the true entropy. Figure \ref{fig:entropy} shows that the larger the model the better the perturb-max bound. 

Both our log-partition bounds as well as our entropy bounds hold in expectation. Thus we evaluate their measure concentration properties, i.e., how many samples are required to converge to their expected value. We evaluate our approach on a $100 \times 100$ spin glass model with $n = 10^4$ variables. The local field parameters $\theta_i$ were drawn uniformly at random from $[-1,1]$ to reflect high signal. To find the perturb-max assignment for such a large model we restrict ourselves to attractive coupling setting; the parameters $\theta_{i,j}$ were drawn uniformly from $[0,c]$, where $c \in [0,4]$ to reflect weak, medium and strong coupling potentials. Throughout our experiments we evaluate the expected value of our bounds with $100$ different samples. We note that both our log-partition and entropy upper bounds have the same gradient with respect to their random perturbations, so their measure concentration properties are the same. In the following we only report the concentration of our entropy bounds; the same concentration occurs for our log-partition bounds. 

Figure \ref{fig:samplemean} shows the error in the sample mean $\frac{1}{M}\sum_{j=1}^M F_j$ as described in Section \ref{sec:bounds}. We do so for three different sample sizes $M=1,5,10$, while $F(\gumset) = \sum_i \gamma_i(x^{\gamma}_i)$ is our entropy bound. The error reduces rapidly as $M$ increases; only $10$ samples are needed to estimate the expectation of the perturb-max random function that consist of $10^4$ random variables $\gamma_i(x_i)$. To test our measure concentration result, that ensures exponential decay, we measure the deviation of the sample mean from its expectation by using $M=1,5,10$ samples. Figure \ref{fig:spin-glass} shows the histogram of the sample mean, i.e., the number of times that the sample mean has error more than $r$ from the true mean. One can see that the decay is indeed exponential for every $M$, and that for larger $M$ the decay is much faster. These show that by understanding the measure concentration properties of MAP perturbations, we can efficiently estimate the mean with high probability, even in very high dimensional spin-glass models. 

\begin{figure}
\centerline{
\begin{tabular}[t]{cccc}
\includegraphics[width=1.8cm]{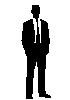} & 
\includegraphics[width=1.8cm]{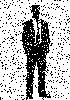} &
\includegraphics[width=1.8cm]{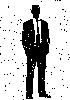} & 
\includegraphics[width=1.8cm]{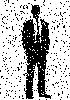} \\ 
model & train / test & ours & SVM-struct 
\end{tabular}
}
\caption{\small \it 
From left to right: (a) Binary $100 \times 70$ image. (b) A representative image in the training set and the test set, where $10\%$ of the pixels are randomly flipped. (c) A  de-noised test image with our method: The test set error is $1.8\%$. (d) A  de-noised test image with SVM-struct: The pixel base error is $2.5\%$.
}    
\label{fig:estimation}
\end{figure}

Lastly, we demonstrate the effectiveness of MAP perturbations in supervised learning. We consider binary image denoising, which is equivalent to learning the parameters of a spin glass model. The training data was composed of ten $100 \times 70$ binary images, consisting of a man in silhouette corrupted by random binary noise, described in Figure \ref{fig:estimation}. Each image $x$ is described by binary local features $\phi_i(x,y_i)$ which equals $1$ if the $i$-th pixel of image $x$ is foreground and $-1$ otherwise. The pairwise features $\phi_{i,j}(y_i,y_i) = 1$ if $y_i = y_j$ and $-1$ otherwise. The goal is to estimate the parameters $\theta_i, \theta_{i,j}$ to de-noise the images. The parameters $\theta_i$ determine the importance of background-foreground observations in pixel $i$, and the parameters $\theta_{i,j}$ determine the coupling nature of pixels $i,j$, namely their attractive or repulsive strength. 

Since we are considering $100 \times 70$ images, there are about $20,000$ parameters to estimate. Conditional random fields cannot be evaluated on this problem, as the partition function cannot be computed for graphs with many cycles. However, the MAP estimate can be efficiently approximated using MPLP. 

Our learning objective function is $\min_\theta \sum_{(x,y) \in S} \frac{1}{|S|} \E_\gamma  [\max\{ \sum_i \theta_i \phi_i(x,y_i) +  \sum_{i,j} \theta_{i,j} \phi_{i,j}(x,y_i,y_j)  + \sum_i \gamma_i(y_i)\}] + \|\theta\|^2$, which serves as an upper bound to the Conditional random fields learning objective. When omitting the perturbations this learning objective is the structured-SVM objective (without label loss).  

We estimated the expected max-perturbation value by evaluating $5$ MAP predictions with random perturbation. We performed gradient decent and stopped either when the gradient step did not improve the objective (decreasing the learning rate by half for $10$ times) or when the algorithm performed $30$ iterations. We compared to structured-SVM by removing the perturbations from our learning algorithm (cf.~\cite{Tsochantaridis06}). Since structured-SVM is a non-smooth program, we used subgradient decent for $10,000$ iterations and learning rate of $10/\sqrt{T}$. For completeness, we note that in our previous evaluation, we ran structured-SVM for $30$ iterations and the results were significantly worse~\cite{HazanJ:12icml}. The running time of the learning algorithms is dominated by the number of MAP evaluations, which is $150$ for learning with random perturbations and $10,000$ with structured-SVM. To evaluate the two learning algorithms, we used MAP prediction on the test data.\footnote{Shpakova and Bach have recently shown that one can improve the test performance by inferring the probabilities of MAP perturbations~\cite{Shpakova16}.} When learning with MAP perturbations, the pixel based error on the test set was $1.8\%$. When learning without perturbations, i.e., with structured-SVMs, the pixel based error on the test set was $2.5\%$.
 
\section{Conclusions \label{sec:conclude}}

High dimensional inference is a key challenge for applying machine learning in real-life applications. While the Gibbs distribution is widely used throughout many areas of research, standard sampling algorithms may be too slow in many cases of interest. In the last few years many optimization algorithms were devised to avoid the computational burden of sampling and instead researchers predicted the most likely (MAP) solution. In this work we explore novel probability models that rely on MAP optimization as their core element. These models measure the robustness of MAP prediction to random shifts of the potential function. We show how to sample from the Gibbs distribution using the expected value of perturb-max operations. We also derive new entropy bounds for perturb-max models that use the expected value of the maximal perturbations. We complete our exploration by investigating the measure concentration of perturb-max value while showing it can be estimated with only a few perturb-max operations. 

The results here can be extended in a number of different directions. In contrast to tree re-weighted or entropy covering bounds, the perturb-max bounds do not have a straightforward tightening scheme. Another direction is to consider the perturb-max model beyond the first moment (expectation). It remains open whether the variance or other related statistics of the perturb-max value can be beneficial for learning, e.g., learning the correlations between data measurements. Understanding the effect of approximate MAP solvers could extend the range of applicability~\cite{Hazan13,Gane14}. A natural extension of these methods is to consider high dimensional learning. Perturb-max models already appear implicitly in risk analysis~\cite{Keshet11} and online learning~\cite{Kalai05}. Novel approaches that consider perturb-max models explicitly~\cite{Tarlow12,Gane14} may derive new learning paradigms for high-dimensional inference.

\appendix

\subsection*{Proof details for Corollary \ref{cor:p-mgf}}

The result in Corollary \ref{cor:p-mgf} follows by taking $C = 4$ in the following lemma.

\begin{lemma}
For any $a > 0$ and $C > 0$, for $\lambda \in [0, 2/a\sqrt{C}]$
	\begin{align}
	\prod_{i=0}^\infty \left(1- \frac{\lambda^2 a^2 C}{4^{i+1}} \right)^{-2^i} \le \frac{2 + \lambda a \sqrt{C}}{2 - \lambda a \sqrt{C}}.
	\end{align}
\label{lem:taylor_bound}
\end{lemma}

\begin{IEEEproof}
To prove this inequality there are three simple steps. First, factor out the first term:
	\begin{align}
	\prod_{i=0}^\infty \left(1- \frac{\lambda^2 a^a C}{4^{i+1}} \right)^{-2^i} = \left(1- \frac{\lambda^2 a^2 C}{4} \right)^{-1}  \prod_{i=1}^\infty \left(1- \frac{\lambda^2 a^2 C}{4^{i+1}} \right)^{-2^i}
	\label{eq:app:first_split}
	\end{align}
and define
	\begin{align}
	V(\lambda) = \prod_{i=1}^\infty \left(1- \frac{\lambda^2 a^2 C}{4^{i+1}} \right)^{-2^i}.
	\label{eq:app:Vdef}
	\end{align}
Next, from the identity
	\begin{align}
	\left(1- \frac{\lambda^2 a^2 C}{4} \right)^{-1} = \frac{4}{(2 +  \lambda a \sqrt{C})(2 - \lambda a \sqrt{C})}.
	\label{eq:app:soloterm}
	\end{align}
If we can show that $\sqrt{ V(\lambda) } < \frac{ 2 + \lambda a \sqrt{C} }{2}$ then the result will follow.

We claim $\sqrt{ V(\lambda) }$ is convex. Note that if $V(\lambda)$ log-convex, then $\sqrt{V(\lambda)}$ is also convex, so it is sufficient to show that $\log V(\lambda)$ is convex. Using the Taylor series expansion $\log( 1 - x ) = - \sum_{j=1}^{\infty} x^j / j$ and switching the order of summation,
	\begin{align}
	\log V(\lambda) &=  -\sum_{i=1}^\infty 2^{i} \log \left(1- \frac{\lambda^2 a^2 C}{4^{i+1}} \right) \\
	&= \sum_{i=1}^\infty 2^{i} \sum_{j=1}^\infty \frac{ (\lambda^{2} a^{2} C)^{j} }{ j \cdot 4^{ji+j}} \\
	&= \sum_{j=1}^\infty \frac{(\lambda^{2} a^{2} C)^{j} }{ j 4^{j}} \sum_{i=1}^\infty \frac{ 2^{i} }{ 2^{ (2 j - 1)i } } \\
	&= \sum_{j=1}^\infty \frac{(\lambda^{2} a^{2} C)^{j} }{ j 4^{j}} \left( \frac{1}{ 1 - 2^{ -(2j - 1) }} - 1 \right) \\
	&=  \sum_{j=1}^\infty \frac{(\lambda^{2} a^{2} C)^{j} }{ j 4^{j}} \left( \frac{1}{ 2^{ 2j - 1} - 1 }  \right).
	\end{align}

Note that the expansion holds only for $\frac{\lambda^2 a^2 C}{4^{i+1}} < 1$ and this bound is tightest for $i=1$. This expansion is the sum of convex functions and hence convex. This means that for $\lambda < \frac{4}{ a \sqrt{C} }$ the function $\sqrt{V(\lambda)}$ is convex.

At $\lambda = \frac{ 2 }{ a \sqrt{C} }$, we have
	\begin{align}
	\log V\left( \frac{ 2 }{ a \sqrt{C} } \right) 
	&= \sum_{j=1}^{\infty} \frac{1}{j} \left( \frac{1}{ 2^{ 2j - 1} - 1 }  \right) \\
	&\le 1 + \sum_{j=2}^{\infty} \frac{1}{ j \cdot 2^{ 2j - 2} }  \\
	&= 1 + \frac{4} \sum_{j=2}^{\infty} \frac{ (1/4)^{j} }{ j } \\
	&= 1 + 4 \left( - \log\left( 1 - \frac{1}{4} \right) - \frac{1}{4} \right) \\
	&= 4 \log \frac{ 4 }{ 3 } \\
	&< \log 4.
	\end{align}
Therefore $V( 2/ a \sqrt{C} ) < 4$.

Since $V(0) = 1$ and $V( 2/ a \sqrt{C} ) < 4$, by convexity, for $\lambda \in [0, 2/a\sqrt{C}]$,
	\begin{align}
	\sqrt{ V(\lambda) } 
	&\le \left(1 - \frac{ \lambda a \sqrt{C} }{2} \right) \sqrt{V(\lambda)} 
		+ \frac{ \lambda a \sqrt{C} }{2}  \sqrt{ V\left( \frac{2}{a\sqrt{C}} \right)  } \\
	&< 1 + \frac{ \lambda a \sqrt{C} }{2} \\
	&= \frac{ 2 + \lambda a \sqrt{C} }{2}.
	\end{align}

Now, considering \eqref{eq:app:first_split} and the terms in \eqref{eq:app:Vdef} and \eqref{eq:app:soloterm}, we have
	\begin{align}
	\prod_{i=0}^\infty \left(1- \frac{\lambda^2 a^a C}{4^{i+1}} \right)^{-2^i}
	&= \frac{4}{(2 +  \lambda a \sqrt{C})(2 - \lambda a \sqrt{C})} V(\lambda) \\
	&< \frac{4}{(2 +  \lambda a \sqrt{C})(2 - \lambda a \sqrt{C})} \cdot \left( \frac{ 2 + \lambda a \sqrt{C} }{2} \right)^2 \\
	&= \frac{ 2 + \lambda a \sqrt{C} }{ 2 - \lambda a \sqrt{C} },
	\end{align}
as desired.
\end{IEEEproof}

\section*{Acknowledgments}

The authors thank the reviewers for their detailed and helpful comments which helped considerably in clarifying the manuscript, Francis Bach and Tatiana Shpakova for helpful discussions, and Associate Editor Constantine Caramanis for his patience and understanding. 

\bibliographystyle{IEEEtran}
\bibliography{logsob}

\end{document}